\documentclass[12pt,journal,draftcls,a4paper,onecolumn]{IEEEtran}

\usepackage{amsfonts,amsmath,amssymb,xcolor,epsfig,graphicx}
\usepackage[normalem]{ulem}
\usepackage{multirow}
\usepackage{dsfont}
\usepackage{amsmath}
\usepackage{amssymb}
\usepackage{amsfonts}
\usepackage{graphicx}
\usepackage{amsmath}
\usepackage[noadjust]{cite}
\usepackage{algorithm}
\usepackage{algorithmic}
\usepackage[all,poly]{xy}
\usepackage{multicol}
\usepackage{accents}
\usepackage{float}
\usepackage{subfigure}

\title{Nonlinear spectral unmixing of hyperspectral images using Gaussian
processes}

\author{Yoann Altmann$^{(1)}$, Nicolas Dobigeon$^{(1)}$, \\
Steve
McLaughlin$^{(2)}$ and Jean-Yves Tourneret$^{(1)}$
\\
\normalsize $^{(1)}$ University of Toulouse,
IRIT/INP-ENSEEIHT/T\'eSA, 31071 Toulouse, France. \\
$^{(2)}$ School of Engineering and Physical Sciences, Heriot-Watt
University of Edinburgh, U.K.\\
\small\texttt{\{Yoann.Altmann,Nicolas,Nicolas.Dobigeon,Jean-Yves.Tourneret\}@enseeiht.fr}\\
\texttt{s.mclaughlin@hw.ac.uk} }

\newcommand{\bg}{\boldsymbol{g}}

\newcommand{\bp}{\boldsymbol{p}}

\newcommand{\bu}{\boldsymbol{u}}

\newcommand{\bv}{\boldsymbol{v}}
\newcommand{\bw}{\boldsymbol{w}}
\newcommand{\bx}{\boldsymbol{x}}

\newcommand{\bz}{\boldsymbol{z}}

\newcommand{\bC}{\boldsymbol{C}}

\newcommand{\bE}{\boldsymbol{E}}

\newcommand{\bK}{\boldsymbol{K}}

\newcommand{\bP}{\boldsymbol{P}}

\newcommand{\bS}{{\boldsymbol S}}

\newcommand{\bU}{\boldsymbol{U}}
\newcommand{\bV}{\boldsymbol{V}}
\newcommand{\bW}{\boldsymbol{W}}
\newcommand{\bX}{{\boldsymbol X}}

\newcommand{\ba}{\boldsymbol{\alpha}}

\newcommand{\bPsi}{{\boldsymbol \Psi}}
\newcommand{\bpsi}{{\boldsymbol \psi}}

\newcommand{\bSigma}{{\boldsymbol \Sigma}}
\newcommand{\bkappa}{{\boldsymbol \kappa}}

\newcommand{\bLambda}{{\boldsymbol \Lambda}}

\newcommand{\blue}{\textcolor{blue} }

\newcommand{\Vpix}[1]{\mathbf{y}_{#1}}

\newcommand{\MATpix}{\mathbf{Y}}

\newcommand{\Vpixels}{\mathbf{y}}

\newcommand{\nbpix}{N}
\newcommand{\nopix}{n}

\newcommand{\nbband}{L}

\newcommand{\nbmat}{R}
\newcommand{\nomat}{r}

\newcommand{\Vmat}[1]{{\mathbf m}_{#1}}

\newcommand{\MATabond}{{\bold A}}

\newcommand{\Vabond}[1]{{\boldsymbol{a}}_{#1}}
\newcommand{\abonds}[1]{a_{#1}}
\newcommand{\Vabonds}{{\boldsymbol{a}}}

\newcommand{\Vnoise}[1]{{\mathbf e}_{#1}}
\newcommand{\Vnoises}[1]{{\boldsymbol{\epsilon}}_{#1}}

\newcommand{\paramvect}{\boldsymbol{\theta}}

\newcommand{\Domain}{\mathcal{D}}

\newcommand{\transp}{^T}

\newcommand{\Ndistr}[1]{\mathcal{N}\left(#1\right)}

\newcommand{\norm}[1]{\left\|#1\right\|}

\newcommand{\Vun}[1]{{\boldsymbol{1}}_{#1}}
\newcommand{\Vzero}{\boldsymbol{0}}
\newcommand{\Id}[1]{\textbf{I}_{#1}}

\newcounter{algo}
\renewcommand{\thealgo}{\arabic{algo}}

\begin{document}
\maketitle

\begin{abstract}
This paper presents an unsupervised algorithm for nonlinear unmixing
of hyperspectral images. The proposed model assumes that the pixel
reflectances result from a nonlinear function of the abundance
vectors associated with the pure spectral components. We assume that
the spectral signatures of the pure components and the nonlinear
function are unknown. The first step of the proposed method consists
of the Bayesian estimation of the abundance vectors for all the
image pixels and the nonlinear function relating the abundance
vectors to the observations. The endmembers are subsequently
estimated using Gaussian process regression. The performance of the
unmixing strategy is evaluated with simulations conducted on
synthetic and real data.
\end{abstract}


\section{Introduction}
Spectral unmixing (SU) is a major issue when analyzing hyperspectral
images. It consists of identifying the macroscopic materials present
in an hyperspectral image and quantifying the proportions of these
materials in the image pixels. Many SU strategies assume that pixel
reflectances are linear combinations of pure component spectra
\cite{Bioucas2012}. The resulting linear mixing model (LMM) has been
widely adopted in the literature and has provided some interesting
results. However, as discussed in \cite{Bioucas2012}, the LMM can be
inappropriate for some hyperspectral images, such as those
containing sand, trees or vegetation areas. Nonlinear mixing models
provide an interesting alternative to overcome the inherent
limitations of the LMM. Nonlinear mixing models recently proposed in
the literature include the bidirectional reflectance-based model of
\cite{Hapke1981} for hyperspectral images including intimate
mixtures, i.e., when the photons are intereacting with all the
materials simultaneously. Such mixtures may occur for instance in
sand or mineral areas. Another class of nonlinear model referred to
as bilinear models have been studied in
\cite{Somers2009,Nascimento2009,Fan2009,Halimi2010} for modeling
scattering effects (mainly observed in vegetation areas). Other more
flexible unmixing techniques have been also proposed to handle wider
classes of nonlinearity, including radial basis function networks
\cite{Guilfoyle2001,Altmann2011IGARSS} post-nonlinear mixing models
\cite{Altmann2012} and kernel-based models
\cite{Broadwater2007,Liu2009,Chen2011,Chen2012}.

Most existing unmixing strategies can be decomposed into two steps
referred to as endmember extraction and abundance estimation.
Endmember identification is usually achieved before estimating the
abundances for all the image pixels. In the last decade, many
endmember extraction algorithms (EEAs) have been developed to
identify the pure spectral components contained in a hyperspectral
image (see \cite{Bioucas2012} for a recent review of these methods).
Most EEAs rely on the LMM, which, as discussed, is inappropriate for
the case of nonlinear mixtures of the endmembers. More recently, an
EEA was proposed in \cite{Heylen2011} to extract endmembers from a
set of nonlinearly mixed pixels. This paper proposes first to
estimate the abundance vectors and to estimate the endmembers during
a second step, using the prediction capacity of Gaussian processes
(GPs). This approach breaks from the usual paradigm of spectral
unmixing. More precisely, this paper considers a kernel-based
approach for nonlinear SU based on a nonlinear dimensionality
reduction using a Gaussian process latent variable model (GPLVM).
The main advantage of GPLVMs is their capacity to accurately model
many different nonlinearities. In this paper, we propose to use a
particular form of kernel based on existing bilinear models, which
allows the proposed unmixing strategy to be accurate when the
underlying mixing model is bilinear. Note that the LMM is a
particular bilinear model. The algorithm proposed herein is
``unsupervised" in the sense that the endmembers contained in the
image and the mixing model are not known. Only the number of
endmembers is assumed to be known. As a consequence, the parameters
to be estimated are the kernel parameters, the endmember spectra and
the abundances for all image pixels.

The paper is organized as follows. Section \ref{sec:mixing_model}
presents the nonlinear mixing model considered in this paper for
hyperspectral image unmixing. Section \ref{sec:GPLVM} introduces the
GPLVM used for latent variable estimation. The constrained GPLVM for
abundance estimation is detailed in Section \ref{sec:scaling}.
Section \ref{sec:endmember_estimation} studies the endmember
estimation procedure using GP regression. Some simulation results
conducted on synthetic data are shown and discussed in Section
\ref{sec:simulations}. Finally, conclusions are drawn in Section
\ref{sec:conclusion}.

\section{Nonlinear mixing model}
\label{sec:mixing_model} Consider a hyperspectral image of $N$
pixels, composed of $R$ endmembers and observed in $L$ spectral
bands. For convenience, the data are assumed to have been previously
centered. The $L$-spectrum
$\Vpixels(n)=[y_{1}(n),\ldots,y_{\nbband}(n)]\transp$ of the $n$th
mixed pixel ($n=1,\ldots,N$) is defined as a transformation of its
corresponding abundance vector
$\Vabonds(n)=[\abonds{1}(n),\ldots,\abonds{R}(n)]\transp$ as follows
\begin{eqnarray}
\label{eq:nonlinear_mapping_abond0}
\Vpixels(\nopix) & = & \bg \left[ \Vabonds(n)\right] + \Vnoise{}(\nopix),\quad n=1,\ldots,N
\end{eqnarray}
where $\bg : \mathbb{R}^R   \rightarrow \mathbb{R}^L$ is a linear or
nonlinear unknown function. The noise vector $\Vnoise{}(\nopix)$ is
an independent, identically distributed (i.i.d.) white Gaussian
noise sequence with variance $\sigma^2$, i.e., $\Vnoise{}(\nopix)
\sim \Ndistr{\Vnoise{}(\nopix)|\Vzero_{\nbband},\sigma^2
\Id{\nbband}}, \nopix =1,\ldots,\nbpix$. Without loss generality,
the nonlinear mapping \eqref{eq:nonlinear_mapping_abond0} from the
abundance space to the observation space can be rewritten
\begin{eqnarray}
\label{eq:nonlinear_mapping_abond1}
\Vpixels(\nopix) & = & \bW_0 \bpsi \left[\Vabonds(n)\right] + \Vnoise{}(\nopix),\quad n=1,\ldots,N
\end{eqnarray}
where $\bpsi: \mathbb{R}^R \rightarrow \mathbb{R}^D$, $\bW_0$ is an
$\nbband \times D$ matrix and the dimension $D$ is the dimension of
the subspace spanned by the transformed abundance vectors $\bpsi
\left[\Vabonds(n)\right], n=1,\ldots,N$. Of course, the performance
of the unmixing strategy relies on the choice of the nonlinear
function $\bpsi$. In this paper, we will use the following
nonlinearity
\begin{eqnarray}
\label{eq:psi}
  \bpsi: & \mathbb{R}^R &   \rightarrow \mathbb{R}^D\nonumber\\
         & \Vabonds     &   \mapsto  \bpsi \left[\Vabonds\right]=\left[ \abonds{1},\ldots,\abonds{R}, \abonds{1}\abonds{2}
            \ldots, \abonds{R-1}\abonds{R}\right]\transp,
\end{eqnarray}
with $D=R(R+1)/2$. The primary motivation for considering this
particular kind of nonlinearity is the fact that the resulting
mixing model is a bilinear model with respect to each abundance
$\abonds{\nomat}, r=1,\ldots,R$. More precisely, this mixing model
extends the generalized bilinear model proposed in \cite{Halimi2010}
and thus the LMM. It is important to note from
\eqref{eq:nonlinear_mapping_abond1} and \eqref{eq:psi} that $\bW_0$
contains the $R$ spectra of the pure components present in the image
and $R(R-1)/2$ interaction spectra between these components. Note
also that the analysis presented in this paper could be applied to
any other nonlinearity $\bpsi$.

Due to physical constraints, the abundance vector
$\Vabonds(n)=[\abonds{1}(n),\ldots,\abonds{R}(n)]\transp$ satisfies
the following positivity and sum-to-one constraints
\begin{equation}
\label{eq:abundancesconst}
\sum_{\nomat=1}^{\nbmat}{\abonds{\nomat}(n)}=1,~~ \abonds{\nomat}(n) \geq
0, \forall \nomat \in \left\lbrace 1,\ldots,\nbmat \right\rbrace.
\end{equation}
Since the nonlinearity $\bpsi$ is fixed, the problem of unsupervised
spectral unmixing is to determine the $L \times D$ spectrum matrix
$\bW_0$, the $N \times R$ abundance matrix
$\MATabond=[\Vabonds(1),\ldots,\Vabonds(N)]\transp$ satisfying
\eqref{eq:nonlinear_mapping_abond1} under the constraints
\eqref{eq:abundancesconst} and the noise variance $\sigma^2$.
Unfortunately, it can be shown that the solution of this constrained
problem is not unique. In the noise-free linear case, it is well
known that the data are contained in a simplex whose vertices are
the endmembers. When estimating the endmembers in the linear case, a
simplex of minimum volume embedding the data is expected.
Equivalently, the estimated abundance vectors are expected to occupy
the largest volume in the simplex defined by
\eqref{eq:abundancesconst}. In a similar fashion to the linear case,
the estimated abundance matrix resulting from an unsupervised
nonlinear SU strategy may not occupy the largest volume in the
simplex defined by \eqref{eq:abundancesconst}. To tackle this
problem, we first propose to relax the positivity constraints for
the elements of the matrix $\MATabond$ and to consider only the
sum-to-one constraint. For ease of understanding, we introduce
$\nbmat \times 1$ vectors satisfying the sum-to-one constraint
\begin{equation}
\label{eq:latent_const}
\sum_{\nomat=1}^{\nbmat}{x_{\nomat}(n)}=1,\quad n=1,\ldots,N
\end{equation}
referred to as \emph{latent variables} and denoted as
$\bx(n)=[x_{1}(n),\ldots,x_{R}(n)]\transp, n=1,\ldots,N$. The
positivity constraint will be handled subsequently by a scaling
procedure discussed in Section \ref{sec:scaling}. The next section
presents the Bayesian model for latent variable estimation using
GPLVMs.

\section{Bayesian model}
\label{sec:GPLVM} 
\begin{figure}[h!]
  \centering
  \includegraphics[width=\columnwidth]{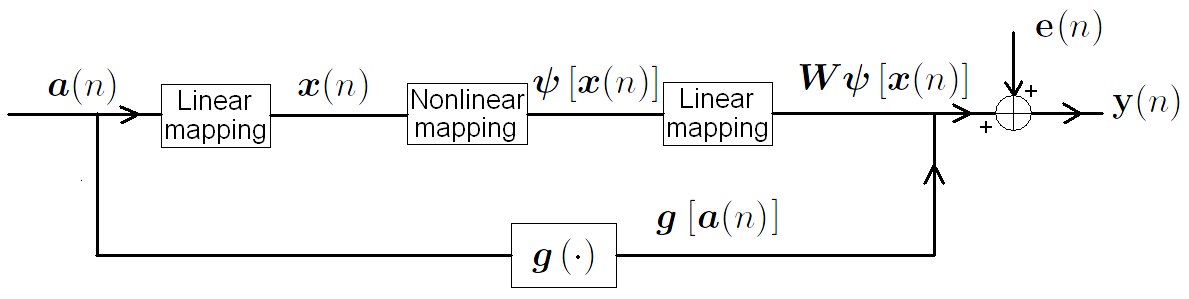}
  \caption{Nonlinear mapping from the abundances vectors to the observed mixed pixels.}
  \label{fig:nonlinear_chain}
\end{figure}

\begin{figure}[h!]
  \centering
  \includegraphics[width=\columnwidth]{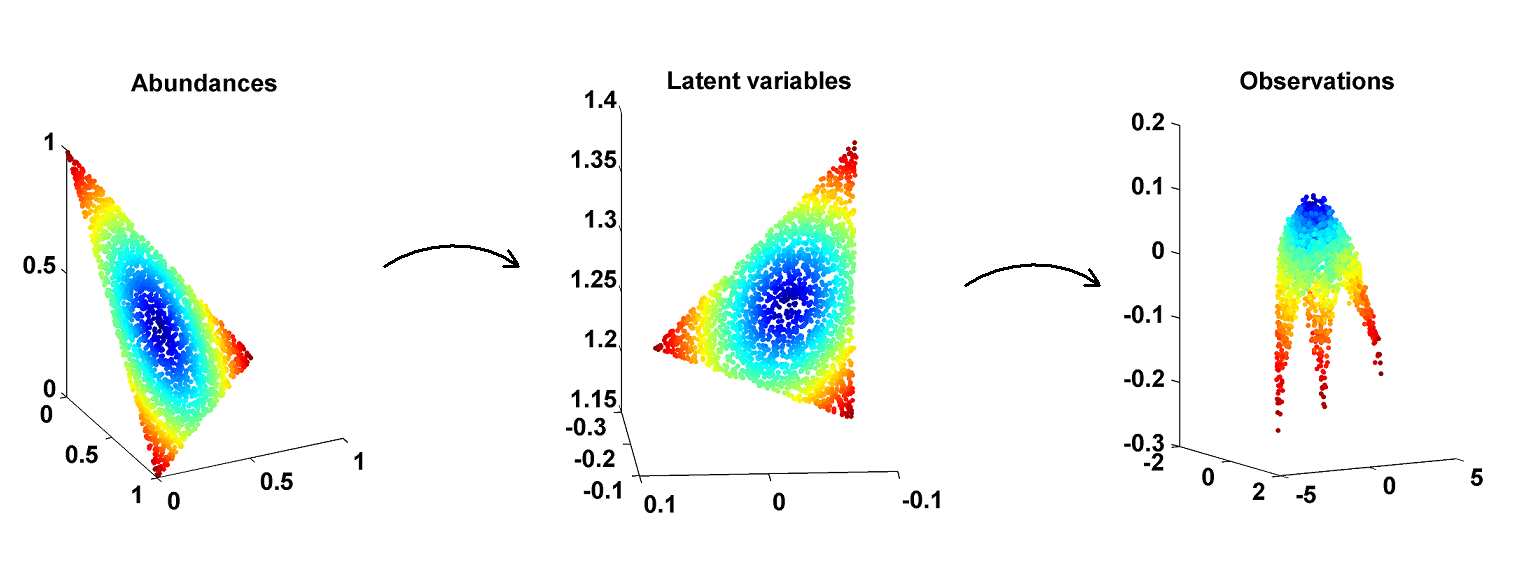}
  \caption{Example of mapping decomposition from the abundance vectors to the observed nonlinearly mixed pixels through the latent variables ($R=3$).}
  \label{fig:mapping_abond}
\end{figure}

GPLVMs \cite{Lawrence2003} are powerful tools for
probabilistic nonlinear dimensionality reduction that rewrite the
nonlinear model \eqref{eq:nonlinear_mapping_abond0} as a nonlinear
mapping from a latent space to the observation space as follows
\begin{eqnarray}
\label{eq:nonlinear_mapping_X}
\Vpixels(\nopix) & = & \bW \bpsi \left[\bx(\nopix)\right] + \Vnoise{}(\nopix),\quad n=1,\ldots,N
\end{eqnarray}
where $\bpsi$ is defined in \eqref{eq:psi},
$\bW=[\bw_{1},\ldots,\bw_{L}]\transp$ is an $\nbband \times D$
matrix with $\bw_{\ell}=[w_{\ell,1},\ldots,w_{\ell,D}]\transp$, and
$D=R(R+1)/2$. Note that from \eqref{eq:nonlinear_mapping_abond1} and
\eqref{eq:nonlinear_mapping_X} the columns of $\bW$ span the same
subspace as the columns of $\bW_0$. Consequently, the columns of
$\bW$ are linear combinations of the spectra of interest, i.e., the
columns of $\bW_0$. Note also that when $\bW$ is full rank, it can
be shown that the latent variables are necessarily linear
combinations of the abundance vectors of interest. Figs.
\ref{fig:nonlinear_chain} and \ref{fig:mapping_abond} illustrate the
mapping from the abundance vectors to the observations that will be
used in this paper. Note that the linear mapping between the
abundances and the latent variables will be explained in detail in
Section \ref{sec:scaling}. For brevity, the $D \times 1$ vectors
$\bpsi \left[\bx(\nopix)\right]$ will be denoted as $\bpsi_{x}(n)$
in the sequel. Assuming independence between the observations, the
statistical properties of the noise lead to the following likelihood
of the $N \times \nbband$ observation matrix
$\MATpix=[\Vpixels(1),\ldots,\Vpixels(N)]\transp$
\begin{eqnarray}
\label{eq:joint_likelihood}
\MATpix|\bW , \bX, \sigma^2 \sim \prod_{n=1}^{N} \Ndistr{\Vpixels(\nopix)|\bW \bpsi_{x}(n),\sigma^2 \Id{\nbband}}
\end{eqnarray}
where $\bX=[\bx(1),\ldots,\bx(N)]\transp$ is the $N \times R$ latent
variable matrix. Note that the likelihood can be rewritten as a
product of Gaussian distributions over the spectral bands as follows
\begin{eqnarray}
\label{eq:joint_likelihood_L}
 \MATpix|\bW , \bX, \sigma^2 \sim \prod_{\ell=1}^{\nbband} \Ndistr{\Vpix{\ell}|\bPsi_{x}\bw_{\ell},\sigma^2 \Id{\nbband}}
\end{eqnarray}
where $\MATpix=[\Vpix{1},\ldots,\Vpix{\nbband}]$ and
$\bPsi_{x}=[\bpsi_{x}(1),\ldots,\bpsi_{x}(N)]\transp$ is an $N
\times D$ matrix. The idea of GPLVMs is to consider $\bW$ as a
nuisance parameter, to assign a Gaussian prior to $\bW$ and to
marginalize the joint likelihood \eqref{eq:joint_likelihood} over
$\bW$, i.e.,
\begin{eqnarray}
\label{eq:marginalization}
f(\MATpix|\bX,\sigma^2) = \int  f(\MATpix|\bW,\bX,\sigma^2)f(\bW) \mathrm{d} \bW
\end{eqnarray}
where $f(\bW)$ is the prior distribution of $\bW$. The estimation of
$\bX$ and $\sigma^2$ can then be achieved by maximizing
\eqref{eq:marginalization} following the maximum likelihood
estimator (MLE) principle. An alternative consists of using an
appropriate prior distribution $f(\bX,\sigma^2)$, assuming prior
independence between $\bW$ and $(\bX,\sigma^2)$, and maximizing the
joint posterior distribution
\begin{eqnarray}
\label{eq:joint_posterior}
f(\bX,\sigma^2|\MATpix) & \propto & \int  f(\MATpix|\bW,\bX,\sigma^2)f(\bW)f(\bX,\sigma^2)\mathrm{d} \bW\nonumber\\
& \propto &  f(\bX,\sigma^2) \int  f(\MATpix|\bW,\bX,\sigma^2)f(\bW) \mathrm{d} \bW \nonumber\\
& \propto &  f(\MATpix|\bX,\sigma^2) f(\bX,\sigma^2)
\end{eqnarray}
with respect to (w.r.t.) $(\bX,\sigma^2)$, yielding the maximum a
posteriori (MAP) estimator of $(\bX,\sigma^2)$. The next paragraph
discusses different possibilities for marginalizing the joint
likelihood \eqref{eq:joint_likelihood_L} w.r.t. $\bW$.

\subsection{Marginalizing $\bW$}
\label{subsec:Marginalization_W} It can be seen from
\eqref{eq:marginalization} that the marginalized likelihood and thus
the associated latent variables depend on the choice of the prior
$f(\bW)$. More precisely, assigning a given prior for $\bW$ favors
particular representations of the data, i.e., particular solutions
for the latent variable matrix $\bX$ maximizing the posterior
\eqref{eq:joint_posterior}. When using GPLVMs for dimensionality
reduction, a classical choice \cite{Lawrence2003} consists of
assigning independent Gaussian priors for $\bw_{1},\ldots,\bw_{L}$,
leading to
\begin{eqnarray}
\label{eq:classical_prior}
f(\bW) = \left(\dfrac{1}{2 \pi}\right)^{\frac{D \nbband}{2}}
 \prod_{\ell=1}^{\nbband} \exp \left[-\dfrac{1}{2} \norm{\bw_{\ell}}^2\right].
\end{eqnarray}
However, this choice can be inappropriate for SU. First, Eq.
\eqref{eq:classical_prior} can be incompatible with the admissible
latent space, constrained by \eqref{eq:latent_const}. Second, the
prior \eqref{eq:classical_prior} assumes the columns of $\bW$
(linear combinations of the spectra of interest) are \emph{a priori}
Gaussian, which is not relevant for real spectra in most
applications. A more sophisticated choice consists of considering
\emph{a priori} correlation between the columns (inter-spectra
correlation) and rows (inter-bands correlation) of $\bW$ using a
structured covariance matrix to be fixed or estimated. In
particular, introducing correlation between close spectral bands is
of particular interest in hyperspectral imagery. Structured
covariance matrices have already been considered in the GP
literature for vector-valued kernels \cite{Bonilla07} (see
\cite{AlvarezRL11arxiv} for a recent review). However, computing the
resulting marginalized likelihood usually requires the estimation of
the structured covariance matrix and the inversion of an $NL \times
NL$ covariance matrix\footnote{See technical report
\cite{AltmannTR2012} for further details.}, which is prohibitive for
SU of hyperspectral images since several hundreds of spectral bands
are usually considered when analyzing real data. Sparse
approximation techniques might be used to reduce this computational
complexity (see \cite{Quinonero-candela05} for a recent review).
However, to our knowledge, these techniques rely on the inversion of
matrices bigger than $N \times N$ matrices. The next section
presents an alternative that only requires the inversion of an $D
\times D$ covariance matrix without any approximation.

\subsection{Subspace identification}
\label{subsec:subspace_identification} It can be seen from
\eqref{eq:nonlinear_mapping_X} that in the noise-free case, the data
belong to a $D$-dimensional subspace that is spanned by the columns
of $\bW$. To reduce the computational complexity induced by the
marginalization of the matrix $\bW$ while considering correlations
between spectral bands, we propose to marginalize a basis of the
subspace spanned by $\bW$ instead of $\bW$ itself. More precisely,
$\bW$ can be decomposed as follows
\begin{eqnarray}
\label{eq:subspace_identification}
\bW= \bP \bU\transp
\end{eqnarray}
where $\bP=[\bp_1,\ldots,\bp_{\nbband}]\transp$ is an $L \times D$
matrix ($\bp_{\ell}$ is $D \times 1$ vector) whose columns are
arbitrary basis vectors of the $D$-dimensional subspace that
contains the subspace spanned by the columns of $\bW$ and
$\bU=[\bu_1,\ldots,\bu_D]\transp$ is a $D \times D$ matrix that
scales the columns of $\bP$. Note that the subspaces spanned by
$\bP$ and $\bW$ are the same when $\bW$ is full rank, resulting in a
full rank matrix $\bU$. The joint likelihood
\eqref{eq:joint_likelihood_L} can be rewritten as
\begin{eqnarray}
\label{eq:joint_likelihood_P}
\MATpix| \bP, \bU, \bX,  \sigma^2 \sim \prod_{\ell=1}^{\nbband} \Ndistr{\Vpix{\ell}|\bC \bp_{\ell},\sigma^2 \Id{\nbband}}
\end{eqnarray}
where $\bC=\bPsi_{x}\bU$ is an $N \times D$ matrix. The proposed
subspace estimation procedure consists of assigning an appropriate
prior distribution to $\bP$ (denoted as $f(\bP)$) and to marginalize
$\bP$ from the joint posterior of interest. It is easier to choose
an informative prior distribution $f(\bP)$ that accounts for
correlation between spectral bands than choosing an informative
$f(\bW)$ since $\bP$ is an arbitrary basis of the subspace spanned
by $\bW$, which can be easily estimated (as will be shown in the
next section).

\subsection{Parameter priors}
GPLVMs construct a smooth mapping from the latent space to the
observation space that preserves dissimilarities
\cite{Lawrence2006}. In the SU context, it means that pixels that
are spectrally different have different latent variables and thus
different abundance vectors. However, preserving local distances is
also interesting: spectrally close pixels are expected to have
similar abundance vectors and thus similar latent variables. Several
approaches have been proposed to preserve similarities, including
back-constraints \cite{Lawrence2006}, dynamical models
\cite{Wang2006} and locally linear embedding (LLE)
\cite{Urtasun2007}. In this paper, we use LLE to assign an
appropriate prior to $\bX$. First, the $K$ nearest neighbors $\{
\Vpixels(j)\}_{j \in \mathcal{\nu}_i}$ of each observation vector
$\Vpixels(i)$ are computed using the Euclidian distance
($\mathcal{\nu}_i$ is the set of integers $j$ such that
$\Vpixels(j)$ is a neighbor of $\Vpixels(i)$). The weight matrix
$\boldsymbol{\Lambda}_{\mathrm{LLE}}=[\lambda_{i,j}]$ of size $N
\times N$ providing the best reconstruction of $\Vpixels(i)$ from
its neighbors is then estimated as
\begin{eqnarray}
 \label{eq:reconstruction_LLE}
\boldsymbol{\Lambda}_{\mathrm{LLE}}= \arg \underset{\boldsymbol{\Lambda}}{\min}
\sum_{i=1}^{N}\norm{\Vpixels(i)-\sum_{j \in \mathcal{\nu}_i}{\lambda_{i,j}
\Vpixels(j)}}^2.
\end{eqnarray}
Note that the solution of \eqref{eq:reconstruction_LLE} is easy to
obtain in closed form since the criterion to optimize is a quadratic
function of $\boldsymbol{\Lambda}$. Note also that the matrix
$\boldsymbol{\Lambda}$ is sparse since each pixel is only described
by its $K$ nearest neighbors. The locally linear patches obtained by
the LLE can then be used to set the following prior for the latent
variable matrix
\begin{eqnarray}
\label{eq:prior_X}
f(\bX | \boldsymbol{\Lambda}_{\mathrm{LLE}},\gamma) & \propto & \exp \left[-\dfrac{\gamma}{2} \sum_{i=1}^{N}\norm{\bx(i)-\sum_{j \in
\mathcal{\nu}_i}{\lambda_{i,j} \bx(j)}}^2\right]\nonumber \\
 & \times & \prod_{n=1}^{N}\Vun{\Domain}\left[\bx(n)\right]
\end{eqnarray}
where $\gamma$ is a fixed hyperparameter to be adjusted and
$\Vun{\Domain}(\cdot)$ is the indicator function over the set
$\Domain$ defined by the constraints \eqref{eq:latent_const}.

In this paper, we propose to assign a prior to $\bP$ using the
standard principal component analysis (PCA) (note again that the
data have been centered). Assuming prior independence between
$\bp_1,\ldots,\bp_{\nbband}$, the following prior is considered for
the matrix $\bP$
\begin{eqnarray}
\label{eq:prior_P}
f\left(\bP|\overline{\bP} ,s^2\right) =\left(\dfrac{1}{2 \pi s^2}\right)^{\frac{N \nbband}{2}}
 \prod_{\ell=1}^{\nbband} \exp \left[-\dfrac{1}{2s^2} \norm{\bp_{\ell}-\bar{\bp}_{\ell}}^2\right]
\end{eqnarray}
where
$\overline{\bP}=[\bar{\bp}_1,\ldots,\bar{\bp}_{\nbband}]\transp$ is
an $L \times D$ projection matrix containing the first $D$
eigenvectors of the sample covariance matrix of the observations
(provided by PCA) and $s^2$ is a dispersion parameter that controls
the dispersion of the prior. Note that the correlation between
spectral bands is implicitly introduced through $\overline{\bP}$.

\begin{figure}[h!]
  \centering
  \includegraphics[width=0.7\columnwidth]{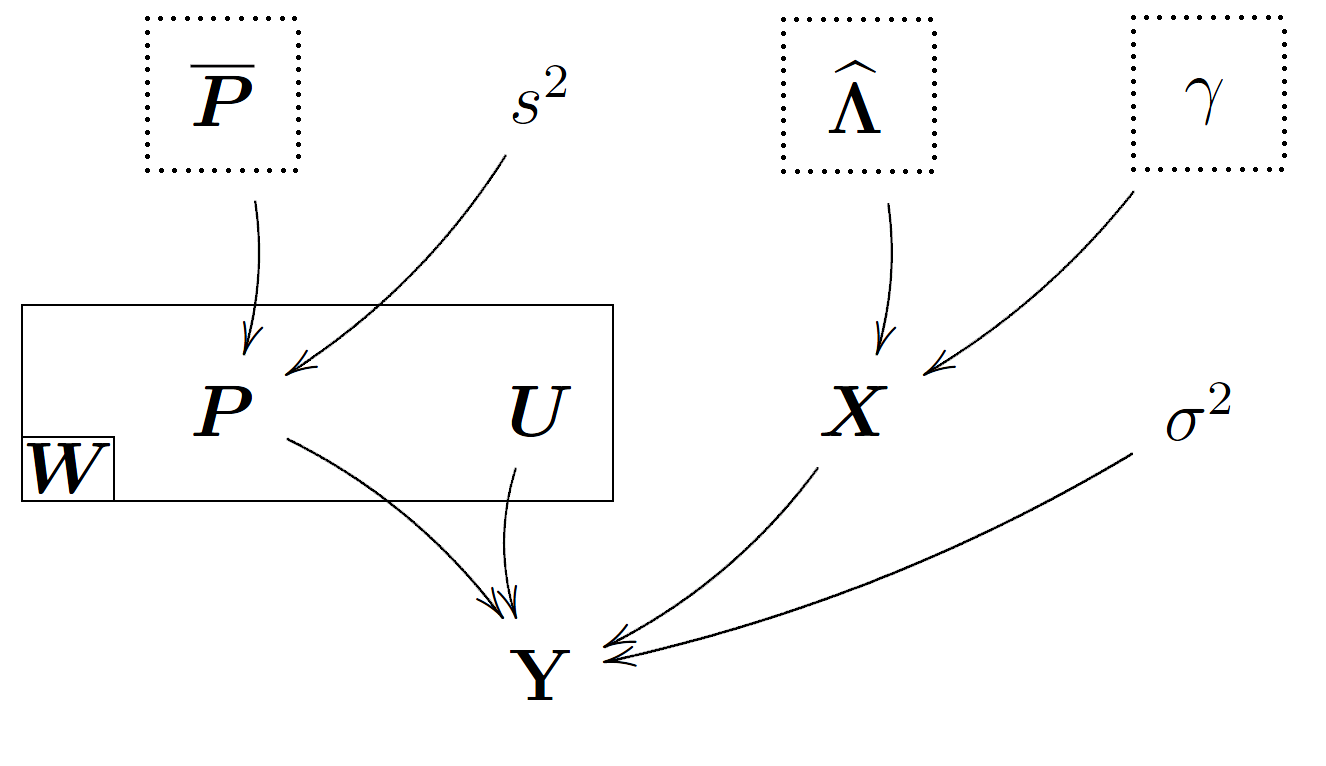}
  \caption{DAG for the parameter priors and hyperpriors (the
fixed parameters appear in dashed boxes).}
  \label{fig:DAG}
\end{figure}

Non-informative priors are assigned to the noise variance $\sigma^2$
and the matrix $\bU$, i.e,
\begin{eqnarray}
\label{eq:sigma2_U_prior}
\begin{array}{ccc}
f(\sigma^2) & \propto & \Vun{(0,\delta_{\sigma^2})}(\sigma^2)\\
f(u_{i,j}) & \propto & \Vun{(-\delta_{\bU},\delta_{\bU})}(u_{i,j})\\
\end{array}
\end{eqnarray}
where the intervals $(0,\delta_{\sigma^2})$ and
$(-\delta_{\bU},\delta_{\bU})$ cover the possible values of the
parameters $\sigma^2$ and $\bU$. Similarly, the following
non-informative prior is assigned to the hyperparameter $s^2$
\begin{eqnarray}
\label{eq:s2_prior}
    f(s^2) & \propto & \Vun{(0,\delta_{s^2})}(s^2)
\end{eqnarray}
where the interval $ (0,\delta_{s^2})$ covers the possible values of
the hyperparameter $s^2$. The resulting directed acyclic graph (DAG)
is depicted in Fig. \ref{fig:DAG}.

\subsection{Marginalized posterior distribution}
\label{subsec:posterior} Assuming prior independence between $\bP$,
$\bX$, $\bU$, $s^2$ and $\sigma^2$, the marginalized posterior
distribution of $\paramvect=(\bX, \bU, s^2, \sigma^2)$ can be
expressed as

\vspace{0.3cm} $f\left(\paramvect | \MATpix,
\bLambda_{\mathrm{LLE}}, \overline{\bP},\gamma \right)$
\vspace{-.1cm}
\begin{eqnarray}
\label{eq:posterior_X}
\vspace{-1.0cm}
 & \propto & f(\paramvect|\bLambda_{\mathrm{LLE}},\gamma) \int f(\MATpix|\bP, \paramvect)  f\left(\bP|\overline{\bP} ,s^2\right)  \mathrm{d} \bP \nonumber\\
 & \propto & f( \MATpix |\paramvect,\overline{\bP}) f(\paramvect|\bLambda_{\mathrm{LLE}},\gamma)
\end{eqnarray}
where $f(\paramvect|\bLambda_{\mathrm{LLE}},\gamma)=
f(\bX|\bLambda_{\mathrm{LLE}},\gamma) f(\bU)f(s^2)f(\sigma^2)$.
Straightforward computations leads to
\begin{eqnarray}
\label{eq:marginalized_likelihood}
f(\MATpix|\paramvect,\overline{\bP}) & = & \int f(\MATpix|\bP, \paramvect)  f\left(\bP|\overline{\bP} ,s^2\right)  \mathrm{d} \bP \nonumber\\
 & \propto & \prod_{\ell=1}^{L} \dfrac{1}{ |\bSigma|^{\frac{1}{2}}}
\exp \left[-\dfrac{1}{2} \bar{\Vpix{}}_{\ell}\transp\bSigma^{-1}\bar{\Vpix{}}_{\ell}\right]\nonumber\\
 & \propto & |\bSigma|^{-\frac{\nbband}{2}} \exp \left[-\dfrac{1}{2} \textrm{tr}(\bSigma^{-1}\bar{\MATpix}\bar{\MATpix}\transp)\right]
\end{eqnarray}
where $\bSigma=s^2\bC\bC\transp + \sigma^2 \Id{N}$,
$\bar{\Vpix{}}_{\ell}=\Vpix{\ell}-\bC\bar{\bp}_{\ell}$ is an $N
\times 1$ vector,
$\bar{\MATpix}=[\bar{\Vpix{}}_{1},\ldots,\bar{\Vpix{}}_{\nbband}]=\MATpix-\bC\overline{\bP}\transp$
is an $N \times L$ matrix and $\textrm{tr}(\cdot)$ denotes the
matrix trace.

Mainly due to the nonlinearity introduced through the nonlinear
mapping, a closed form expression for the parameters maximizing the
joint posterior distribution \eqref{eq:posterior_X} is impossible to
obtain. We propose to use a scaled conjugate gradient (SCG) method
to maximize the marginalized log-posterior. To ensure the sum-to-one
constraint for $\bX$, the following arbitrary reparametrization
\begin{equation}
x_{R}(n)=1 - \sum_{r=1}^{R-1}{x_{r}(n)},\quad n=1,\ldots,N \nonumber
\end{equation}
is used and the marginalized posterior distribution is optimized
w.r.t. the first $(R-1)$ columns of $\bX$ denoted $\bX_{\backslash
R}$. The partial derivatives of the log-posterior w.r.t.
$\bX_{\backslash R}, \bU, s^2$ and $\sigma^2$ are obtained using
partial derivatives w.r.t. $\bSigma$ and $\bar{\MATpix}$ and the
classical chain rules (see technical report \cite{AltmannTR2012} for
further details). The resulting latent variable estimation procedure
is referred to as locally linear GPLVM (LL-GPLVM).

Note that the marginalized likelihood reduces to the product of $L$
independent Gaussian probability density functions since
\begin{eqnarray}
\label{eq:marginalize_likelihood_L}
\Vpix{\ell}|\bar{\bp}_{\ell},\bU,\bX,\sigma^2,s^2 \sim \Ndistr{\bC\bar{\bp}_{\ell}, s^2\bC\bC\transp + \sigma^2 \Id{N}}
\end{eqnarray}
and $\ell=1,\ldots,\nbband$. Note also that the covariance matrix
$\bSigma=s^2\bC\bC\transp + \sigma^2 \Id{N}$ is related to the
covariance matrix of the $2$nd order polynomial kernel \cite[p.
89]{Rasmussen2005}. More precisely, the proposed nonlinear mapping
corresponds to a particular polynomial kernel whose metric is
induced by the matrix $\bU$. Finally, note that the evaluation of
the marginalized likelihood \eqref{eq:marginalized_likelihood} only
requires the inversion of the $N \times N$ covariance matrix
$\bSigma$. It can been seen from the following Woodbury matrix
identity \cite{brookes05}
\begin{eqnarray}
\bSigma^{-1}=\sigma^{-2}\left[\Id{N} - \bC\left(\sigma^{2}s^{-2}\Id{D} + \bC\transp\bC \right)^{-1}\bC\transp\right]
\end{eqnarray}
that the computation of $\bSigma^{-1}$ mainly relies on the
inversion of a $D \times D$ matrix. Similarly, the computation of
$|\bSigma|=1/|\bSigma^{-1}|$ mainly consists of computing the
determinant of a $D \times D$ matrix, which reduces the
computational cost when compared to the structured covariance matrix
based approach presented in Section \ref{subsec:Marginalization_W}.

\subsection{Estimation of $\bP$}
\label{subsec:estimation_P} Let us denote as
$\hat{\paramvect}=(\widehat{\bX}, \widehat{\bU}, \hat{s}^2,
\hat{\sigma}^2)$ the maximum a posteriori (MAP) estimator of
$\paramvect=(\bX, \bU, s^2, \sigma^2)$ obtained by maximizing
$\eqref{eq:posterior_X}$. Using the likelihood
\eqref{eq:joint_likelihood_P}, the prior distribution
\eqref{eq:prior_P} and Bayes' rule, we obtain the posterior
distribution of $\bP$ conditioned upon $\paramvect$, i.e.,
\begin{eqnarray}
\label{eq:posterior_P}
\bP|\MATpix,\paramvect,\overline{\bP} \sim \prod_{\ell=1}^{\nbband} \Ndistr{\bp_{\ell}|\hat{\bp}_{\ell}\,\bS}
\end{eqnarray}
where $\bS^{-1}=\sigma^{-2}\bC\transp\bC + s^{-2}\Id{D}$ and
$\hat{\bp}_{\ell}=\bS(\bC\transp\Vpixels_{\ell}-\bar{\bp}_{\ell})$.
Since the conditional posterior distribution of $\bP$ is the product
of $L$ independent Gaussian distributions, the MAP estimator of
$\bP$ conditioned upon $\hat{\paramvect}$ is given by
\begin{equation}
\label{eq:MAP_P}
\widehat{\bP}=\left(\MATpix\transp\widehat{\bC}-\overline{\bP}\right)\widehat{\bS}
\end{equation}
where
$\widehat{\bS}^{-1}=\hat{\sigma}^{-2}\widehat{\bC}\transp\widehat{\bC}
+ \hat{s}^{-2}\Id{D}$,
$\widehat{\bC}=\widehat{\bPsi}_{x}\widehat{\bU}$,
$\widehat{\bPsi}_{x}=[\bpsi_{\hat{x}}(1),\ldots,\bpsi_{\hat{x}}(N)]\transp$
and $\widehat{\bX}=[\hat{\bx}(1),\ldots,\hat{\bx}(N)]\transp$. The
next section studies a scaling procedure that estimates the
abundance matrix using the estimated latent variables resulting from
the maximization of \eqref{eq:posterior_X}.

\section{Scaling procedure}
\label{sec:scaling} The optimization procedure presented in Section
\ref{subsec:posterior} provides a set of latent variables that
represent the data but can differ from the abundance vectors of
interest. Consider $\widehat{\bX}=[\widehat{\bX}_{\backslash R},
\Vun{N}-\widehat{\bX}_{\backslash R}\Vun{R-1}]$ obtained after
maximization of the posterior \eqref{eq:posterior_X}. The purpose of
this section is to estimate an $N \times R$ abundance matrix
$\MATabond=[\Vabonds(1),\ldots,\Vabonds(N)]\transp$  such that
\begin{eqnarray}
\label{eq:linear_mapping}
\widehat{\bX}_{\backslash R}=\MATabond \bV_{R-1}\transp + \bE
\end{eqnarray}
where $\Vabonds(1),\ldots,\Vabonds(N)$ occupy the maximal volume in
the simplex defined by \eqref{eq:abundancesconst},
$\bV_{R-1}=[\bv_{1},\ldots,\bv_{R}]$ is an $(R-1) \times R$ matrix
and $\bE$ is an $N \times (R-1)$ standard i.i.d Gaussian noise
matrix which models the scaling errors. Since $\widehat{\bX}$
satisfy the sum-to-one constraint \eqref{eq:latent_const},
estimating the relation between $\widehat{\bX}_{\backslash R}$ and
$\MATabond$ is equivalent to estimate the relation between
$\widehat{\bX}$ and $\MATabond$. However, when considering the
mapping between $\widehat{\bX}$ and $\MATabond$, non-isotropic noise
has to be considered since the rows of $\widehat{\bX}$ and
$\MATabond$ satisfy the sum-to-one constraint, i.e., they belong to
the same $(R-1)$-dimensional subspace.

Eq. \eqref{eq:linear_mapping} corresponds to a LMM whose noisy
observations are the rows of $\widehat{\bX}_{\backslash R}$. Since
$\MATabond$ is assumed to occupy the largest volume in the simplex
defined by \eqref{eq:abundancesconst}, the columns of $\bV_{R-1}$
are the vertices of the simplex of minimum volume that contains
$\widehat{\bX}_{\backslash R}$. As a consequence, it seems
reasonable to use a linear unmixing strategy for the set of vectors
$\widehat{\bx}_{\backslash R}(1),\ldots,\hat{\bx}_{\backslash R}(N)$
to estimate $\MATabond$ and $\bV_{R-1}$. In this paper, we propose
to estimate jointly $\MATabond$ and $\bV_{R-1}$ using the Bayesian
algorithm presented in \cite{Dobigeon2009} for unsupervised SU
assuming the LMM. Note that the algorithm in \cite{Dobigeon2009}
assumed positivity constraints for the estimated endmembers. Since
these constraints for $\bV_{R-1}$ are unjustified, the original
algorithm has slightly been modified by removing the truncations in
the projected endmember priors (see \cite{Dobigeon2009} for
details). Once the estimator
$(\widehat{\MATabond},\widehat{\bV}_{R-1})$ of
$(\MATabond,\bV_{R-1})$ has been obtained by the proposed scaling
procedure, the resulting constrained latent variables denoted as
$\widehat{\bX}^{(c)}=[\hat{\bx}^{(c)}(1),\ldots,\hat{\bx}^{(c)}(N)]\transp$
are defined as follows
\begin{eqnarray}
\label{eq:constrained_X}
\widehat{\bX}^{(c)} & = & \begin{bmatrix}
\widehat{\bX}_{\backslash R}^{(c)}, & \Vun{N}-\widehat{\bX}_{\backslash R}^{(c)}\Vun{R-1}
\end{bmatrix}
\end{eqnarray}
with $\widehat{\bX}_{\backslash
R}^{(c)}=\widehat{\MATabond}\widehat{\bV}_{R-1}\transp$. Using the
sum-to-one constraint $\widehat{\MATabond}\Vun{R}=\Vun{N}$, we
obtain
\begin{eqnarray}
\widehat{\bX}^{(c)} & = & \begin{bmatrix}
\widehat{\MATabond}\widehat{\bV}_{R-1}\transp, & \widehat{\MATabond}\Vun{R}- \widehat{\MATabond}\widehat{\bV}_{R-1}\transp\Vun{R-1}\transp
\end{bmatrix} \nonumber\\
 & = &  \widehat{\MATabond}
\begin{bmatrix}
\widehat{\bV}_{R-1}\transp, & \Vun{R}-\widehat{\bV}_{R-1}\transp\Vun{R-1}
\end{bmatrix}\\
 & = & \widehat{\MATabond} \widehat{\bV}_{R}\transp\nonumber
\end{eqnarray}
where $\widehat{\bV}_{R}=[\widehat{\bV}_{R-1}\transp,~~
\Vun{R}-\widehat{\bV}_{R-1}\transp\Vun{R-1}]\transp$ is an $R \times
R$ matrix. The final abundance estimation procedure, including the
LL-GPLVM presented in Section \ref{sec:GPLVM} and the scaling
procedure investigated in this section is referred to as fully
constrained LL-GPVLM (FCLL-GPLVM) (a detailed algorithm is available
in \cite{AltmannTR2012}). Once the final abundance matrix
$\widehat{\MATabond}$ and the matrix $\widehat{\bV}_{R}$ have been
estimated, we propose an endmember extraction procedure based on GP
regression. This method is discussed in the next section.

\section{Gaussian process regression}
\label{sec:endmember_estimation} Endmember estimation is one of the
main issues in SU. Most of the existing EEAs intend to estimate the
endmembers from the data, i.e., selecting the most pure pixels in
the observed image \cite{Chaudhry2006,Winter1999,Nascimento2005}.
However, these approaches can be inefficient when the image does not
contain enough pure pixels. Some other EEAs based on the
minimization of the volume containing the data (such as the minimum
volume simplex analysis \cite{Li2008igarss}) can mitigate the
absence of pure pixels in the image. This section studies a new
endmember estimation strategy based on GP regression for nonlinear
mixtures. This strategy can be used even when the scene does not
contain pure pixels. It assumes that all the image abundances have
been estimated using the algorithm described in Section
\ref{sec:scaling}. Consider the set of pixels $\left\lbrace
\Vpixels(n)\right\rbrace_{n=1,\ldots,N}$ and corresponding estimated
abundance vectors $\left\lbrace
\hat{\Vabonds}(n)\right\rbrace_{n=1,\ldots,N}$. GP regression first
allows the nonlinear mapping $\bg(\cdot)$ in
\eqref{eq:nonlinear_mapping_abond0} (from the abundance space to the
observation space) to be estimated. The estimated mapping is denoted
as $\hat{\bg}(\cdot)$. Then, it is possible to use the prediction
capacity of GPs to predict the spectrum
$\hat{\bg}(\boldsymbol{\alpha})$ corresponding to any new abundance
vector $\boldsymbol{\alpha}$. In particular, the predicted spectra
associated with pure pixels, i.e., the endmembers, correspond to
abundance vectors that are the vertices of the simplex defined by
\eqref{eq:abundancesconst}. This section provides more details about
GP prediction for endmember estimation.

It can be seen from the marginalized likelihood
\eqref{eq:marginalized_likelihood} that $f(\MATpix|\bX,
\overline{\bP}, \bU, s^2, \sigma^2)$ is the product of $\nbband$
independent GPs associated with each spectral band of the data space
\eqref{eq:marginalize_likelihood_L}. Looking carefully at the
covariance matrix of $\Vpix{\ell}$ (i.e., to
$\bSigma=s^2\bC\bC\transp + \sigma^2 \Id{N}$), we can write
\begin{eqnarray}
\label{eq:decomposition}
\Vpix{\ell}=\bz_{\ell} + \Vnoises{\ell}
\end{eqnarray}
where $ \Vnoises{\ell}$ is the $N \times 1$ white Gaussian noise
vector associated with the $\ell$th spectral band (having covariance
matrix $\sigma^2 \Id{N}$) and\footnote{Note that all known
conditional parameters have been omitted for brevity.}
\begin{eqnarray}
\label{eq:GP_prior}
\bz_{\ell} \sim \Ndistr{\bz_{\ell}| \bPsi_{x}\bU\bar{\bp}_{\ell},\bK}
\end{eqnarray}
with $\bK=s^2\bPsi_{x}\bU\bU\transp\bPsi_{x}\transp$ the $N \times
N$ covariance matrix of $\bz_{\ell}$. The $N \times 1$ vector
$\bz_{\ell}$ is referred to as hidden vector associated with the
observation $\Vpix{\ell}$. Consider now an $\nbband \times 1$ test
data with hidden vector $\bz^*=[z_1^*,...,z_L^*]^T$, abundance
vector $\ba^*=[\alpha_{1}^*,...,\alpha_{\nbmat}^*]^T$ and
$\bpsi_{x}^*=\bpsi\left[\bV_{R}\ba^* \right]$. We assume that the
test data share the same statistical properties as the training data
$\Vpix{1},...,\Vpix{L}$ in the sense that
 $[\bz_{\ell}^T,z_{\ell}^*]$ is a Gaussian vector
such that
\begin{eqnarray}
\label{eq:joint_bf_f}
 \begin{bmatrix}
  \bz_{\ell}\\
  z_{\ell}^*
 \end{bmatrix} \sim \Ndistr{\begin{bmatrix}
  \bz_{\ell}\\
  z_{\ell}^*
 \end{bmatrix} \bigg| \begin{bmatrix}
  \bPsi_{x}\bU\bar{\bp}_{\ell}\\
  \bpsi_{x}^{*T}\bU\bar{\bp}_{\ell}
 \end{bmatrix}
  , \begin{bmatrix}
  \bK & \bkappa(\ba^*) \\
  \bkappa(\ba^*)\transp & \sigma_{\ba^*}^2
 \end{bmatrix}
 }
\end{eqnarray}
where $\sigma_{\ba^*}^2=s^2 \bpsi_{x}^{*T}\bU\bU\transp \bpsi_{x}^*$
is the variance of $z_{\ell}^*$ and $\bkappa(\ba^*)$ contains the
covariances between the training inputs and the test inputs, i.e.,
\begin{eqnarray}
\bkappa(\ba^*)= s^2 \bpsi_{x}^{*T}\bU\bU\transp\bPsi_{x}.
\end{eqnarray}
Straightforward computations leads to
\begin{eqnarray}
\label{eq:prediction_endmember}
z_{\ell}^*|\Vpix{\ell} \sim \Ndistr{z_{\ell}^* | \mu_{\ell},s_{l}^2}
\end{eqnarray}
with
\begin{eqnarray}
    \begin{array}{ccc}
    \mu_{\ell} & = & \bpsi_{x}^{*T}\bU\bar{\bp}_{\ell} +
    \bkappa(\ba^*)\transp(\bK + \sigma^2\Id{N})^{-1}(\Vpix{\ell}-
    \bPsi_{x}\bU\hat{\bp}_{\ell}) \nonumber \\
    s_{l}^2 & = & \sigma_{\ba^*}^2-\bkappa(\ba^*)\transp(\bK +
    \sigma^2\Id{N})^{-1}\bkappa(\ba^*).
    \end{array}
\end{eqnarray}
Since the posterior distribution \eqref{eq:prediction_endmember} is
Gaussian, the MAP and MMSE estimators of $\bz^*$ equal the posterior
mean $\boldsymbol{\mu}=(\mu_1,...,\mu_L)^T$.

In order to estimate the endmembers, we propose to replace the
parameters $\bX, \bU, s^2$ and $\sigma^2$ by their estimates
$\widehat{\bX}^{(c)}, \widehat{\bU}, \hat{s}^2$ and $\hat{\sigma}^2$
and to compute the estimated hidden vectors associated with the
abundance vectors
$\ba^*=[\boldsymbol{0}_{r-1}^T,1,\boldsymbol{0}_{R-r}^T]^T$ for
$r=1,...,R$. For each value of $r$, the $r$th estimated hidden
vector will be the $r$th estimated endmember\footnote{Note that the
estimated endmembers are centered since the data have previously
been centered. The actual endmembers can be obtained by adding the
empirical mean to the estimated endmembers.}. Indeed, for the LMM
and the bilinear models considered in this paper, the endmembers are
obtained by setting
$\ba^*=[\boldsymbol{0}_{r-1}^T,1,\boldsymbol{0}_{R-r}^T]^T$ in the
model \eqref{eq:nonlinear_mapping_abond1} relating the observations
to the abundances. Note that the proposed endmember estimation
procedure provides the posterior distribution of each endmember via
\eqref{eq:prediction_endmember} which can be used to derive
confidence intervals for the estimates. The next section presents
some simulation results obtained for synthetic and real data.

\section{Simulations}
\label{sec:simulations}
\subsection{Synthetic data}
\begin{figure}[h!]
  \centering
  \includegraphics[width=\columnwidth]{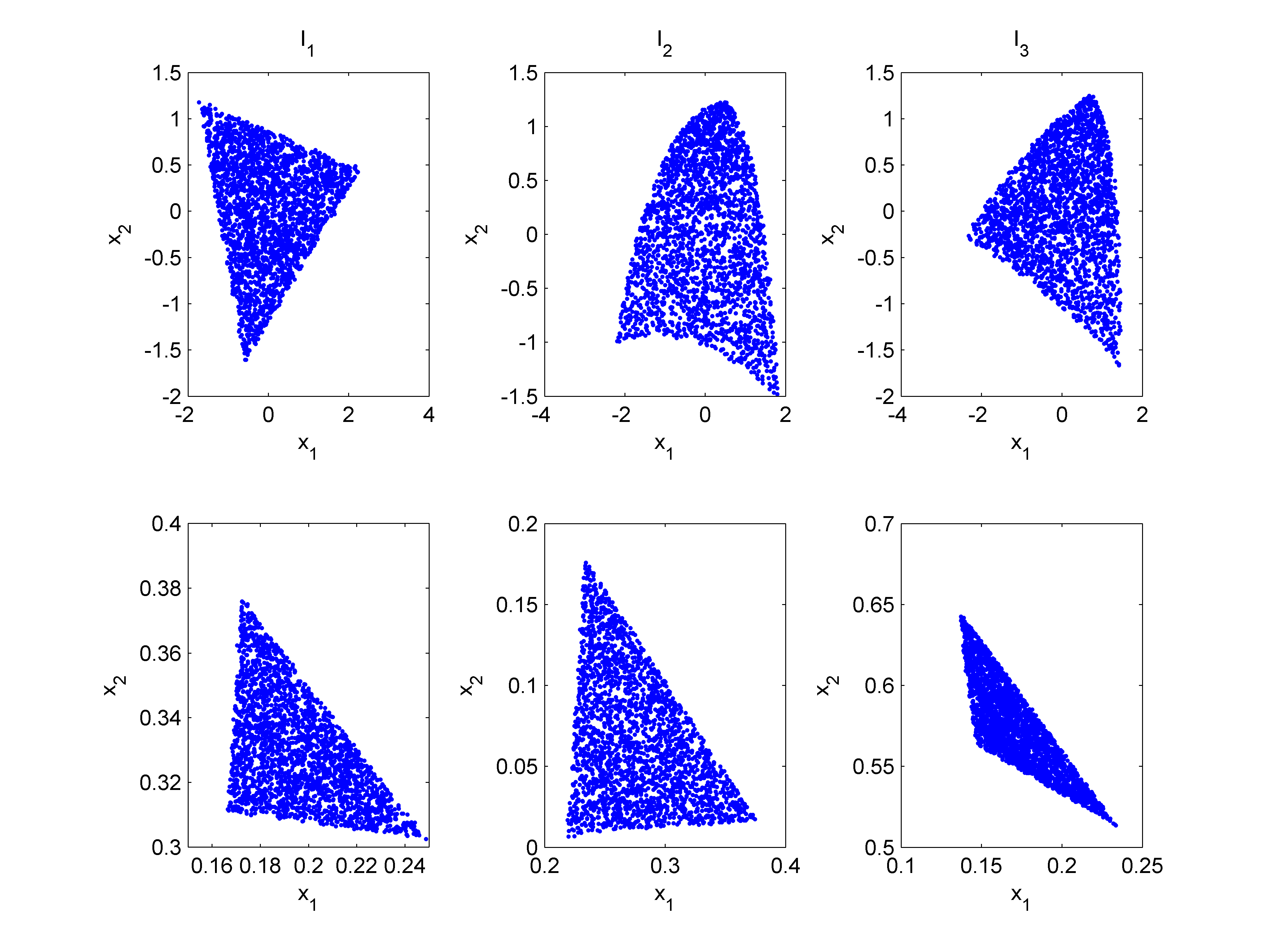}
  \caption{Top: Representation of the $N=2500$ pixels (dots) using the first two principal components
  provided by the standard PCA for the three synthetic images $I_1$ to $I_3$.
  Bottom: Representation using the latent variables estimated by the LL-GPLVM for the three synthetic images $I_1$ to $I_3$.}
\label{fig:PCA}
\end{figure}

\begin{figure}[h!]
\begin{minipage}[b]{1.0\linewidth}
  \centering

\includegraphics[width=0.45\linewidth]{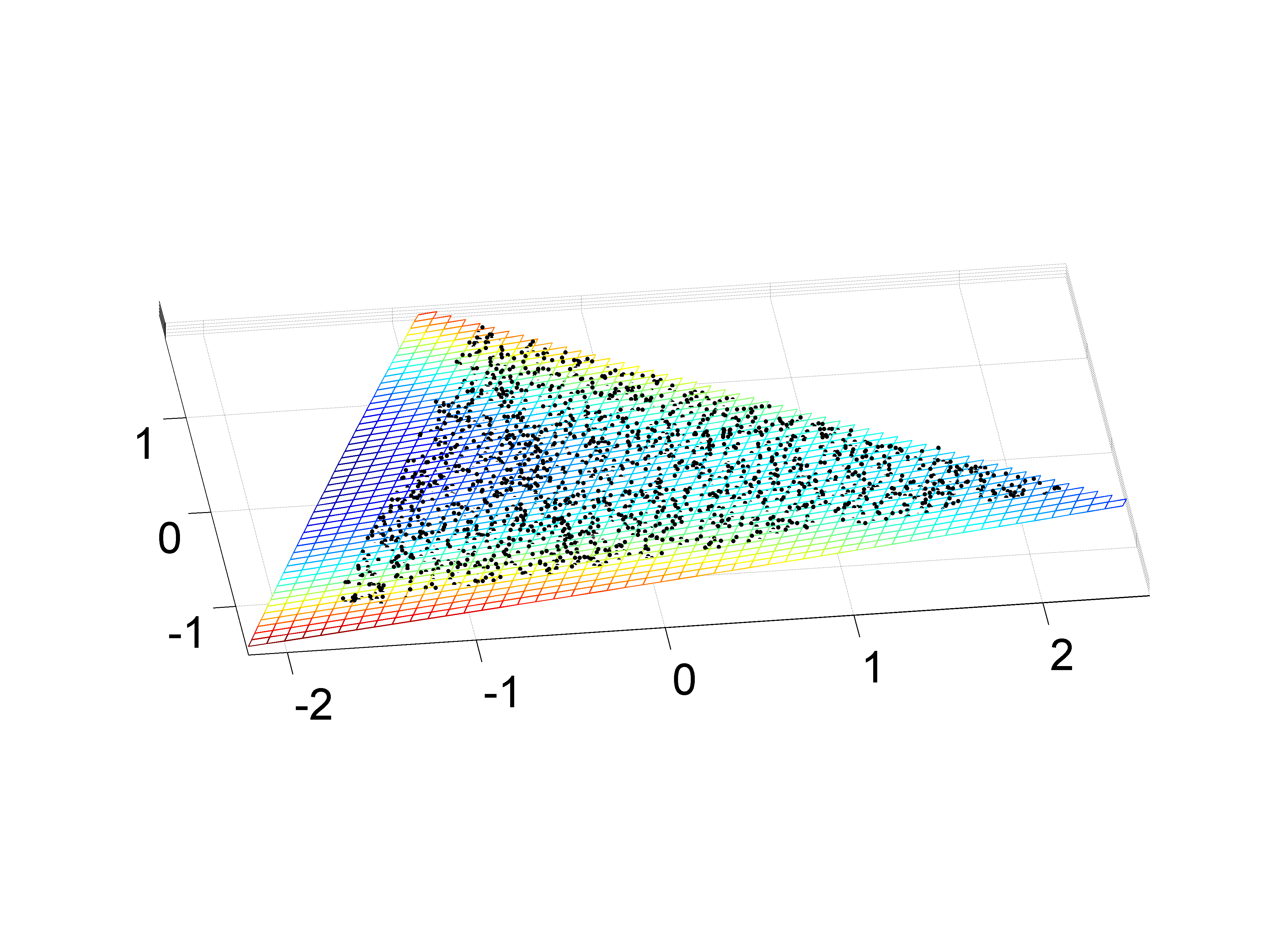}
  \centerline{(a) $I_1$ (LMM)}\medskip
\end{minipage}
\begin{minipage}[b]{0.48\linewidth}
  \centering
\includegraphics[width=\linewidth]{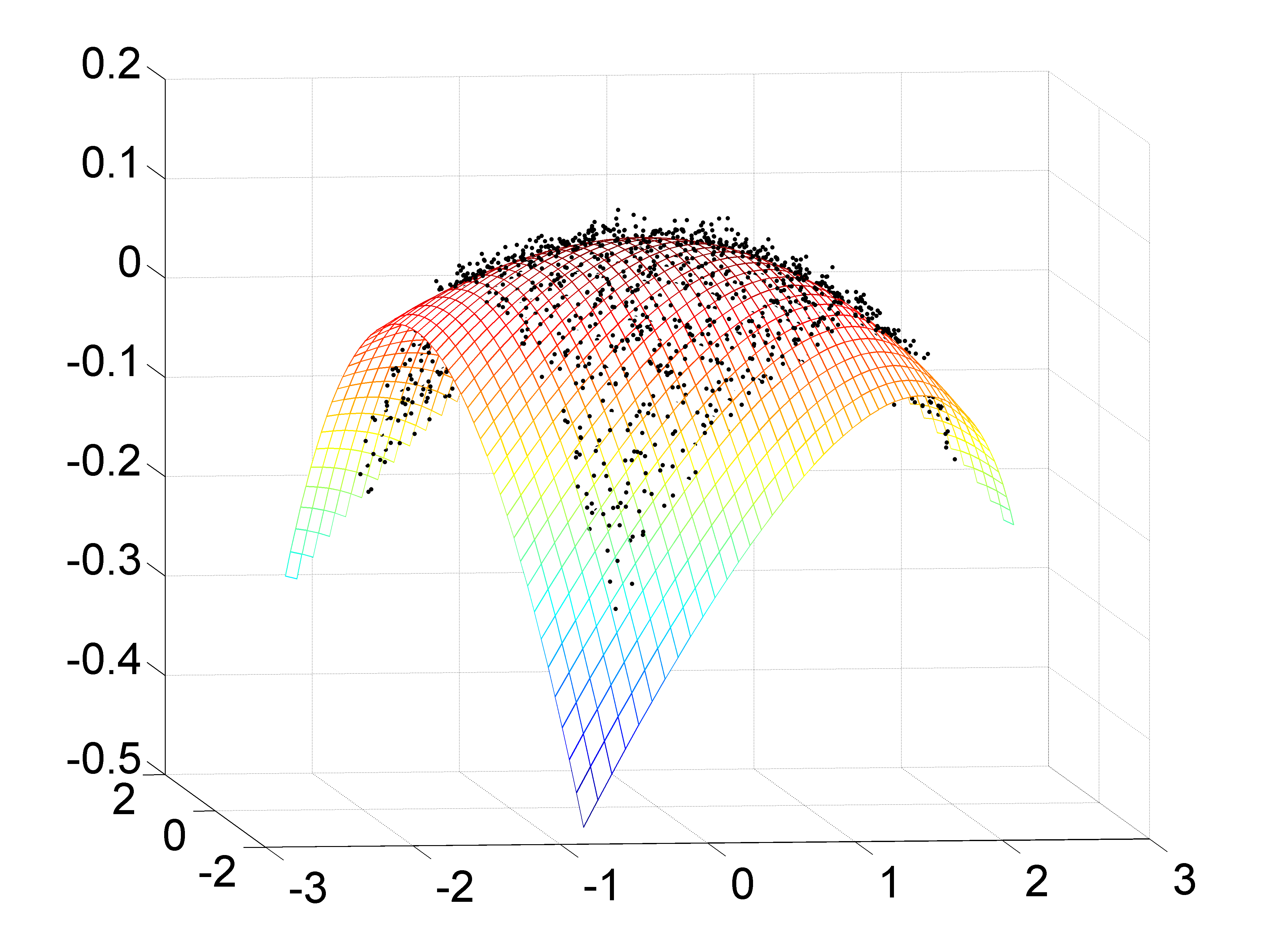}
  \centerline{(b) $I_2$ (FM)}\medskip
\end{minipage}
\hfill
\begin{minipage}[b]{0.48\linewidth}
  \centering
\includegraphics[width=\linewidth]{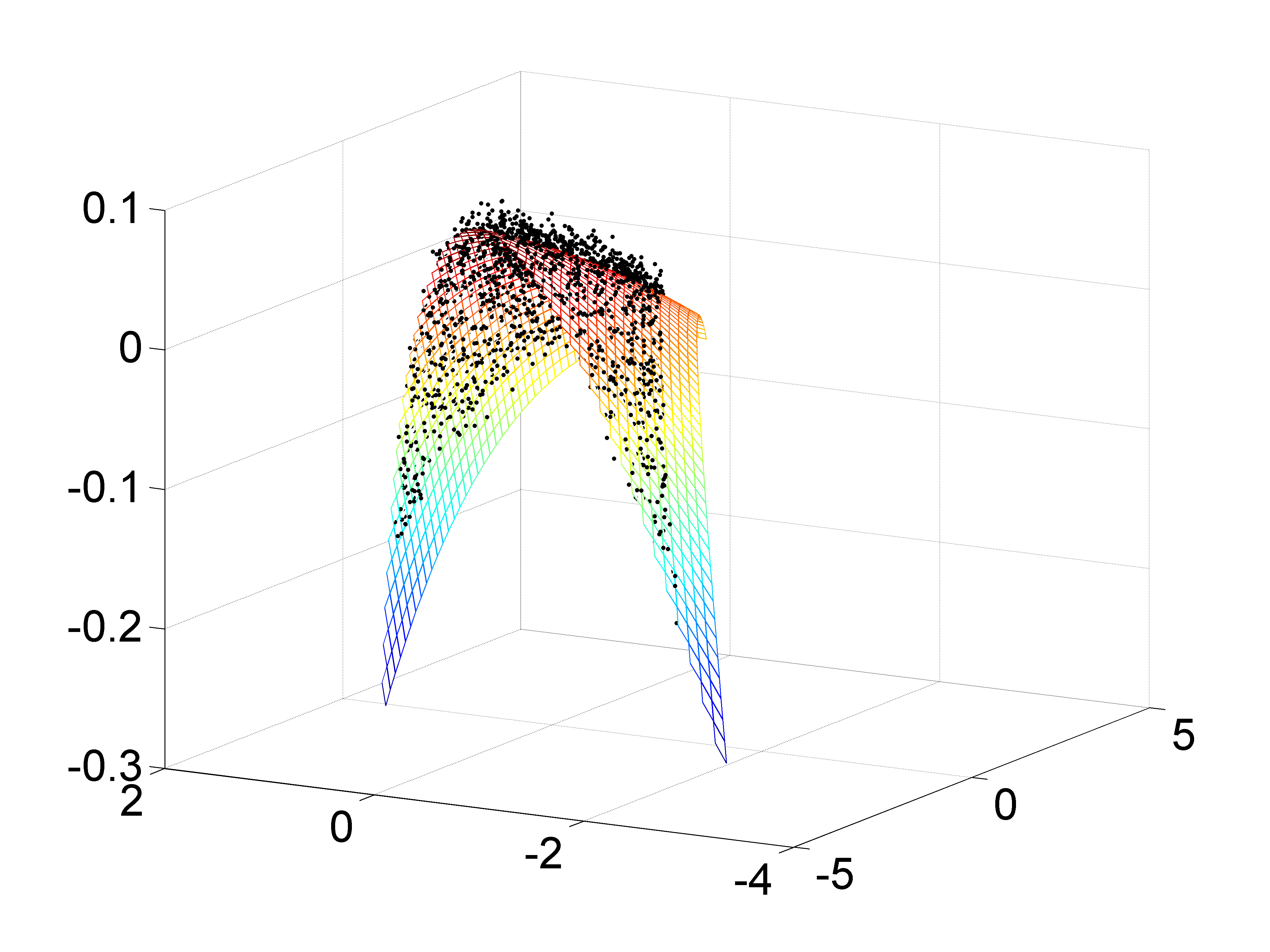}
  \centerline{(c) $I_3$ (GBM)}\medskip
\end{minipage}
\vspace{-0.3cm} \caption{Visualization of the $N=2500$ pixels (black
dots) of $I_1$, $I_2$ and $I_3$  using the 3 axis provided by the
PCA procedure. The colored surface is the manifold identified by the
LL-GPLVM.} \label{fig:estimated_manifolds}
\end{figure}

The performance of the proposed GPLVM for dimensionality reduction
is first evaluated on three synthetic images of $\nbpix=2500$
pixels. The $\nbmat = 3$ endmembers contained in these images have
been extracted from the spectral libraries provided with the ENVI
software \cite{ENVImanual2003} (i.e., green grass, olive green paint
and galvanized steel metal). The first image $I_1$ has been
generated according to the linear mixing model (LMM). The second
image $I_2$ is distributed according to the bilinear mixing model
introduced in \cite{Fan2009}, referred to as the ``Fan model'' (FM).
The third image $I_3$ has been generated according to the
generalized bilinear model (GBM) studied in \cite{Halimi2010} with
the following nonlinearity parameters
\begin{equation}
\gamma_{1,2}=0.9, \quad \gamma_{1,3}=0.5,
    \quad \gamma_{2,3}=0.3. \nonumber
\end{equation}
The abundance vectors $\Vabond{\nopix}, \nopix=1,\ldots,N$ have been
randomly generated according to a uniform distribution on the
admissible set defined by the positivity and sum-to-one constraints
\eqref{eq:abundancesconst}. The noise variance has been fixed to
$\sigma^2=10^{-4}$, which corresponds to a signal-to-noise ratio
$\mathrm{SNR} \approx 30$dB which corresponds to the worst case for
current spectrometers. The hyperparameter $\gamma$ of the latent
variable prior \eqref{eq:prior_X} has been fixed to $\gamma=10^3$
and the number of neighbors for the LLE is $K=R$ for all the results
presented in this paper. The quality of dimensionality reduction of
the GPLVM can be measured by the average reconstruction error (ARE)
defined as
\begin{eqnarray}
\label{eq:RE}
    \textrm{ARE}= \sqrt{\dfrac{1}{\nbband \nbpix}\sum_{n=1}^{N}\norm{\hat{\Vpix{}}_{n}- \Vpix{n}}^2}
\end{eqnarray}
where $\Vpix{n}$ is the $n$th observed pixel and $\hat{\Vpix{}}_{n}$
its estimate. For the LL-GPLVM, the $n$th estimated pixel is given
by $\hat{\Vpix{}}_{n}=\widehat{\bP}\widehat{\bU}\transp\bpsi
\left[\hat{\bx}(n)\right]$ where $\widehat{\bP}$ is estimated using
\eqref{eq:MAP_P}. Table \ref{tab:ARE_synth} compares the AREs
obtained by the proposed LL-GPLVM and the projection onto the first
$(R-1)$ principal vectors provided by the principal component
analysis (PCA). The proposed LL-GPLVM slightly outperforms PCA for
nonlinear mixtures in term of ARE. More precisely, the AREs of the
LL-GPLVM mainly consist of the noise errors ($\sigma^2=10^{-4}$),
whereas model errors are added when applying PCA to nonlinear
mixtures. Fig. \ref{fig:PCA} compares the latent variables obtained
after maximization of \eqref{eq:marginalized_likelihood} for the
three images $I_1$ to $I_3$ with the projections obtained by
projecting the data onto the $R-1$ principal vectors provided by
PCA. Note that only $R-1$ dimensions are needed to represent the
latent variables (because of the sum-to-one constraint). From this
figure, it can be seen that the latent variables of the LL-GPLVM
describe a noisy simplex for the three images. It is not the case
when using PCA for the nonlinear images. Fig.
\ref{fig:estimated_manifolds} shows the manifolds estimated by the
LL-GPLVM for the three images $I_1$ to $I_3$. This figure shows that
the proposed LL-GPLVM can model the manifolds associated with the
image pixels with good accuracy.
\begin{table}[h!]
\renewcommand{\arraystretch}{1.2}
\begin{footnotesize}
\begin{center}
\caption{AREs: synthetic images.\label{tab:ARE_synth}}
\begin{tabular}{|c|c|c|c|c|c|c|}
\cline{2-7}
\multicolumn{1}{c|}{} & \multicolumn{6}{c|}{ARE ($\times 10^{-2}$)} \\
\cline{2-7}
\multicolumn{1}{c|}{}               & $I_1$ & $I_2$ & $I_3$ & $I_1^*$ & $I_2^*$ & $I_3^*$\\
\hline
\multicolumn{1}{|c|}{PCA}           & \textbf{\blue{0.99}} & 1.08 & 1.04 & \textbf{\blue{1.00}} & 1.06  & 1.03 \\
\hline
\multicolumn{1}{|c|}{LL-GPLVM}       & \textbf{\blue{0.99}} & \textbf{\blue{0.99}} & \textbf{\blue{1.00}} & \textbf{\blue{1.00}} & \textbf{\blue{1.00}}  & \textbf{\blue{0.99}}\\
\hline
\hline
\multicolumn{1}{|c|}{SU}            & 1.00 & 1.13 & 1.06 & 1.14 & 1.57 & 1.12\\
\hline
\multicolumn{1}{|c|}{FCLL-GPLVM}       & \textbf{\blue{0.99}} & \textbf{\blue{0.99}} & \textbf{\blue{1.00}} & \textbf{\blue{1.00}} & \textbf{\blue{1.00}} & \textbf{\blue{0.99}}\\
\hline
\end{tabular}
\end{center}
\end{footnotesize}
\vspace{-0.4cm}
\end{table}
The quality of unmixing procedures can also be measured by comparing
the estimated and actual abundances using the root normalized mean
square error (RNMSE) defined by
\begin{eqnarray}
\label{eq:RMSE}
    \textrm{RNMSE}= \sqrt{\dfrac{1}{\nbpix \nbmat}\sum_{\nopix=1}^{\nbpix}
    {\norm{\hat{\Vabonds}_{\nopix} - \Vabond{\nopix}}^2}}
\end{eqnarray}
where $\Vabond{\nopix}$ is the $n$th actual abundance vector and
$\hat{\Vabonds}_{\nopix}$ its estimate. Table \ref{tab:RMSE_synth}
compares the RNMSEs obtained with different unmixing strategies. The
endmembers have been estimated by the VCA algorithm in all
simulations. The algorithms used for abundance estimation are the
FCLS algorithm proposed in \cite{Heinz2001} for $I_1$, the LS method
proposed in \cite{Fan2009} for $I_2$ and the gradient-based method
proposed in \cite{Halimi2010} for $I_3$. These procedures are
referred to as ``SU'' in the table. These strategies are compared
with the proposed FCLL-GPLVM. As mentioned above, the Bayesian
algorithm for joint estimation of $\MATabond$ and $\bV$ under
positivity and sum-to-one constraints for $\MATabond$ (introduced in
\cite{Dobigeon2009}) is used in this paper for the scaling step. It
can be seen that the proposed FCLL-GPLVM is general enough to
accurately approximate the considered mixing models since it
provides the best results in term of abundance estimation.

\begin{table}[h!]
\renewcommand{\arraystretch}{1.2}
\begin{footnotesize}
\begin{center}
\caption{RNMSEs: synthetic images.\label{tab:RMSE_synth}}
\begin{tabular}{|c|c|c|c|c|c|c|}
\cline{2-7}
\multicolumn{1}{c|}{} & \multicolumn{6}{c|}{RNMSE ($\times 10^{-3}$)} \\
\cline{2-7}
\multicolumn{1}{c|}{}               & $I_1$ & $I_2$ & $I_3$ & $I_1^*$ & $I_2^*$ & $I_3^*$\\
\hline
\multicolumn{1}{|c|}{SU}            & 5.7 & 7.4 & 22.7  & 49.3 & 86.6 & 47.8  \\
\hline
\multicolumn{1}{|c|}{FCLL-GPLVM}       & \textbf{\blue{3.9}} & \textbf{\blue{4.2}} & \textbf{\blue{5.4}} & \textbf{\blue{4.8}} & \textbf{\blue{7.2}} & \textbf{\blue{7.5}}\\
\hline
\end{tabular}
\end{center}
\end{footnotesize}
\vspace{-0.4cm}
\end{table}

The quality of reconstruction of the unmixing procedure is also
evaluated by the ARE. For the FCLL-GPLVM, the $n$th reconstructed
pixel $\hat{\Vpix{}}_{n}$ is given by
$\hat{\Vpix{}}_{n}=\widehat{\bP}\widehat{\bU}\transp\bpsi
\left[\hat{\bx}^{(c)}(n)\right]$. Table \ref{tab:ARE_synth} also
shows the AREs corresponding to the different unmixing strategies.
The proposed FCLL-GPLVM outperforms the other strategies in term of
ARE for these images.

Finally, the performance of the FCLL-GPLVM for endmember estimation
is evaluated by comparing the estimated endmembers with the actual
spectra. The quality of endmember estimation is evaluated by the
spectral angle mapper (SAM) defined as
\begin{eqnarray}
\label{eq:SAM}
    \textrm{SAM}= \textrm{arccos}\left(\dfrac{\left \langle \hat{\Vmat{r}}, \Vmat{r}\right \rangle}{\norm{\hat{\Vmat{r}}} \norm{\Vmat{r}}} \right)
\end{eqnarray}
where $\Vmat{r}$ is the $r$th actual endmember and $\hat{\Vmat{r}}$
its estimate. Table \ref{tab:SAM_synth} compares the SAMs obtained
for each endmember using the VCA algorithm, the nonlinear EEA
presented in \cite{Heylen2011} (referred to as ``Heylen'') and the
FCLL-GPLVM for the three images $I_1$ to $I_3$. These results show
that the FCLL-GPLVM provides accurate endmember estimates for both
linear and nonlinear mixtures.
\begin{table}[h!]
\renewcommand{\arraystretch}{1.2}
\begin{footnotesize}
\begin{center}
\caption{SAMs ($\times 10^{-2}$): synthetic
images.\label{tab:SAM_synth}}
\begin{tabular}{|c|c|c|c|c|}
\cline{3-5}
\multicolumn{2}{c|}{} & VCA & Heylen & FCLL-GPLVM\\
\hline
\multirow{3}{*}{$I_1$} & $\Vmat{1}$ & \textbf{\blue{0.43}} & 1.94 & 0.52\\
\cline{2-5}
                       & $\Vmat{2}$ & \textbf{\blue{0.22}} & 0.66 & 0.86\\
\cline{2-5}
                       & $\Vmat{3}$ & 0.22 & 0.78 & \textbf{\blue{0.15}}\\
\hline
\multirow{3}{*}{$I_2$} & $\Vmat{1}$ & 1.62 & 0.75 & \textbf{\blue{0.33}}\\
\cline{2-5}
                       & $\Vmat{2}$ & 2.08 & 1.69 & \textbf{\blue{0.53}}\\
\cline{2-5}
                       & $\Vmat{3}$ & 1.15 & 0.42 & \textbf{\blue{0.34}}\\
\hline
\multirow{3}{*}{$I_3$} & $\Vmat{1}$ & 1.91 & 1.80 & \textbf{\blue{0.44}}\\
\cline{2-5}
                       & $\Vmat{2}$ & 1.36 & 0.86 & \textbf{\blue{0.58}}\\
\cline{2-5}
                       & $\Vmat{3}$ & 0.88 & 1.38 & \textbf{\blue{0.30}}\\
\hline
\end{tabular}
\end{center}
\end{footnotesize}
\vspace{-0.4cm}
\end{table}

\begin{figure}[h!]
  \centering
  \includegraphics[width=\columnwidth]{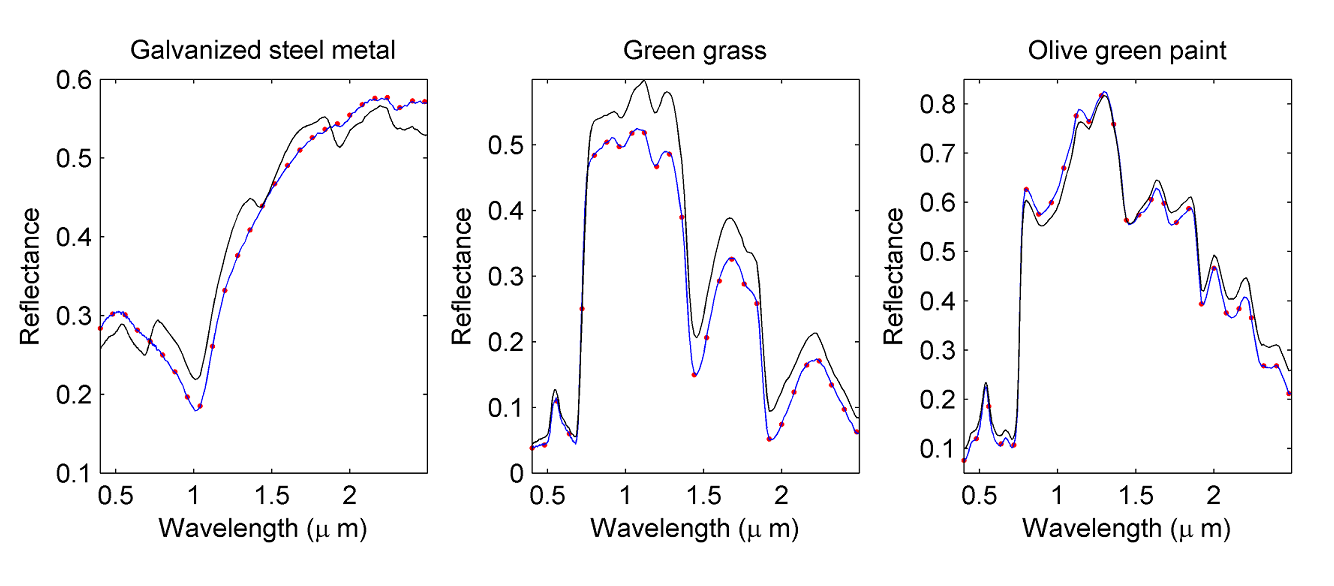}
  \caption{Actual endmembers (red
dots) and endmembers estimated by the FCLL-GPLVM (blue lines) and VCA (black line) for the image $I_2^*$.}
\label{fig:FM_endmembers}
\end{figure}

Finally, the performance of the proposed unmixing algorithm is
tested in scenarios where pure pixels are not present in the
observed scene. More precisely, the simulation parameters remain the
same for the three images $I_1$ to $I_3$ except for the $N=2500$
abundance vectors, that are drawn from a uniform distribution in the
following set
\begin{equation}
\left\lbrace \Vabonds \big| \sum_{\nomat=1}^{\nbmat}{\abonds{\nomat}}=1,
~~ 0.9 \geq \abonds{\nomat}(n) \geq
0, \forall \nomat \in \left\lbrace 1,\ldots,\nbmat \right\rbrace \right\rbrace.
\end{equation}
The three resulting images are denoted as $I_1^*$, $I_2^*$ and
$I_3^*$. Table \ref{tab:ARE_synth} shows that the absence of pure
pixels does not change the AREs significantly when they are compared
with those obtained with the images $I_1$ to $I_3$. Moreover,
FCLL-GPLVM is more robust to the absence of pure pixels than the
different SU methods. The good performance of FCLL-GPVLM is due in
part to the scaling procedure. Table \ref{tab:RMSE_synth} shows that
the performance of the FCLL-GPLVM in term of RNMSE is not degraded
significantly when there is no pure pixel in the image (note that
the situation is different when the endmembers are estimated using
VCA). Table \ref{tab:SAM_synth_truncated} shows the performance of
the FCLL-GPLVM for endmember estimation when there are no pure
pixels in the image. The results of the FCLL-GPLVM do not change
significantly when they are compared with those obtained with images
$I_1$ to $I_3$, which is not the case for the two other EEAs. The
accuracy of the endmember estimation is illustrated in Fig.
\ref{fig:FM_endmembers} which compares the endmembers estimated by
the FCLL-GPLVM (blue lines) to the actual endmember (red dots) and
the VCA estimates (black line) for the image $I_2^*$.

The next section presents simulation results obtained for real data.

\begin{table}[h!]
\renewcommand{\arraystretch}{1.2}
\begin{footnotesize}
\begin{center}
\caption{SAMs ($\times 10^{-2}$): synthetic
images.\label{tab:SAM_synth_truncated}}
\begin{tabular}{|c|c|c|c|c|}
\cline{3-5}
\multicolumn{2}{c|}{} & VCA & Heylen & FCLL-GPLVM\\
\hline
\multirow{3}{*}{$I_1^*$} & $\Vmat{1}$ & 2.87& 6.38& \textbf{\blue{0.38}}\\
\cline{2-5}
                       & $\Vmat{2}$ & 2.15 & 11.11 & \textbf{\blue{1.30}}\\
\cline{2-5}
                       & $\Vmat{3}$ & 2.10 & 2.62 & \textbf{\blue{0.24}}\\
\hline
\multirow{3}{*}{$I_2^*$} & $\Vmat{1}$ & 5.22 & 7.53 & \textbf{\blue{0.67}}\\
\cline{2-5}
                       & $\Vmat{2}$ & 8.02 & 9.59 & \textbf{\blue{1.46}}\\
\cline{2-5}
                       & $\Vmat{3}$ & 7.10 & 2.48 & \textbf{\blue{0.53}}\\
\hline
\multirow{3}{*}{$I_3^*$} & $\Vmat{1}$ & 6.89 & 6.59 & \textbf{\blue{0.61}}\\
\cline{2-5}
                       & $\Vmat{2}$ & 6.03 & 5.95 & \textbf{\blue{1.75}}\\
\cline{2-5}
                       & $\Vmat{3}$ & 3.73 & 2.36 & \textbf{\blue{0.48}}\\
\hline
\end{tabular}
\end{center}
\end{footnotesize}
\vspace{-0.4cm}
\end{table}

\subsection{Real data}
\begin{figure}[h!]
  \centering
  \includegraphics[width=\columnwidth]{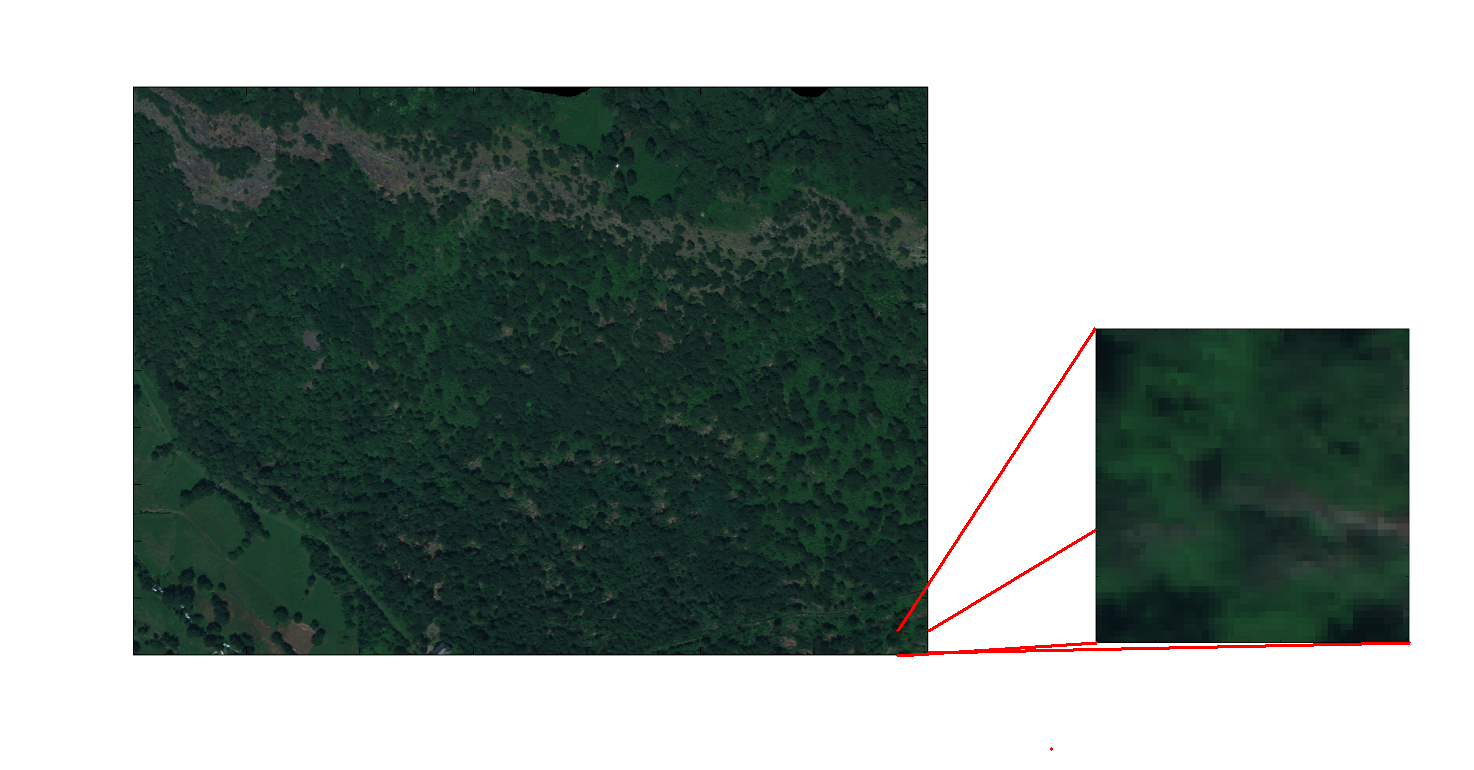}
  \caption{Real hyperspectral data: Madonna data acquired by the Hyspex hyperspectral scanner
   over Villelongue, France (left) and the region of interest shown in true colors (right).}
\label{fig:Madonna_big}
\end{figure}
\begin{figure}[h!]
\begin{minipage}[b]{1.0\linewidth}
  \centering
\includegraphics[width=0.45\linewidth]{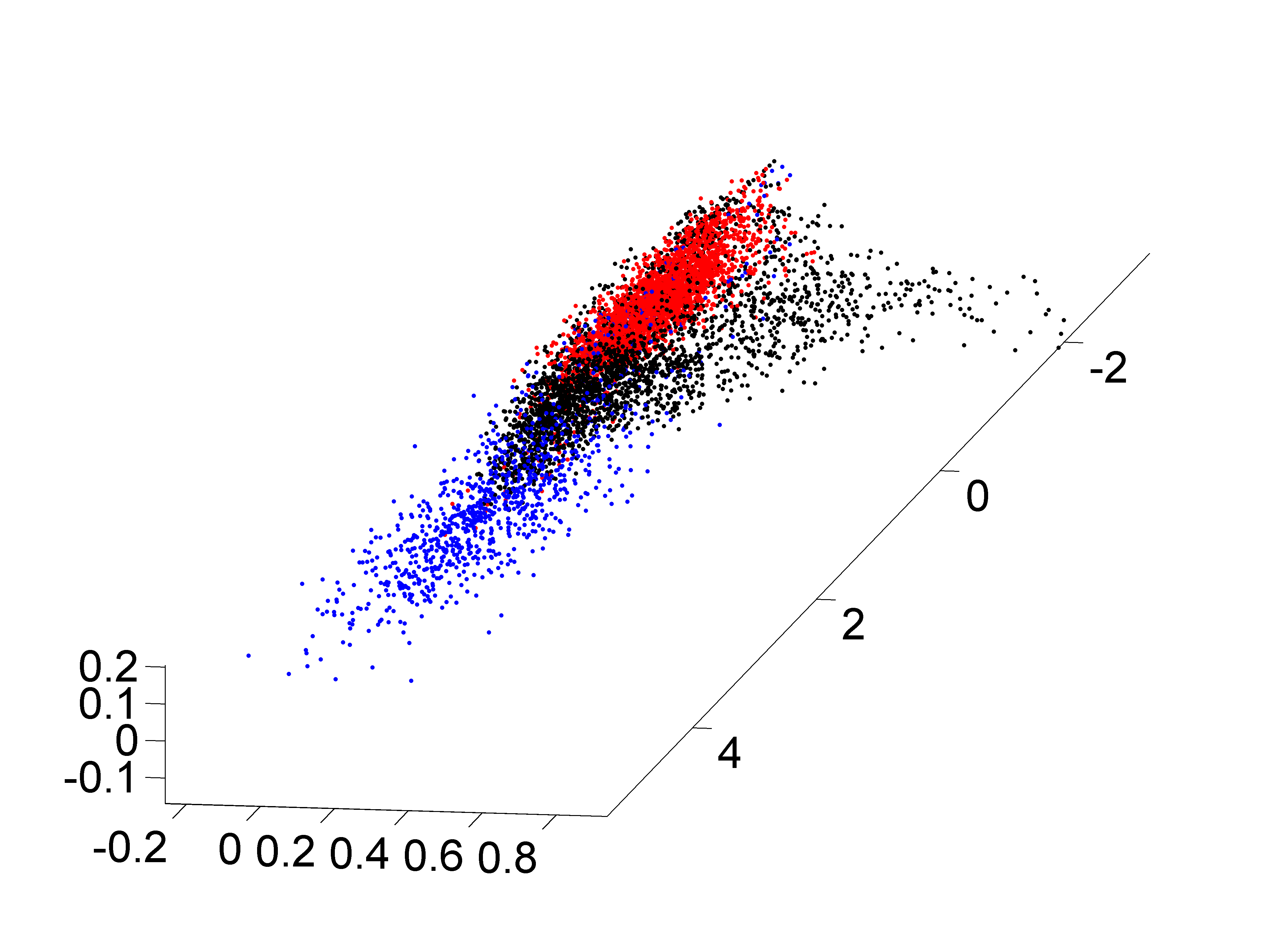}
  \centerline{(a)}\medskip
\end{minipage}
\begin{minipage}[b]{0.48\linewidth}
  \centering
\includegraphics[width=\linewidth]{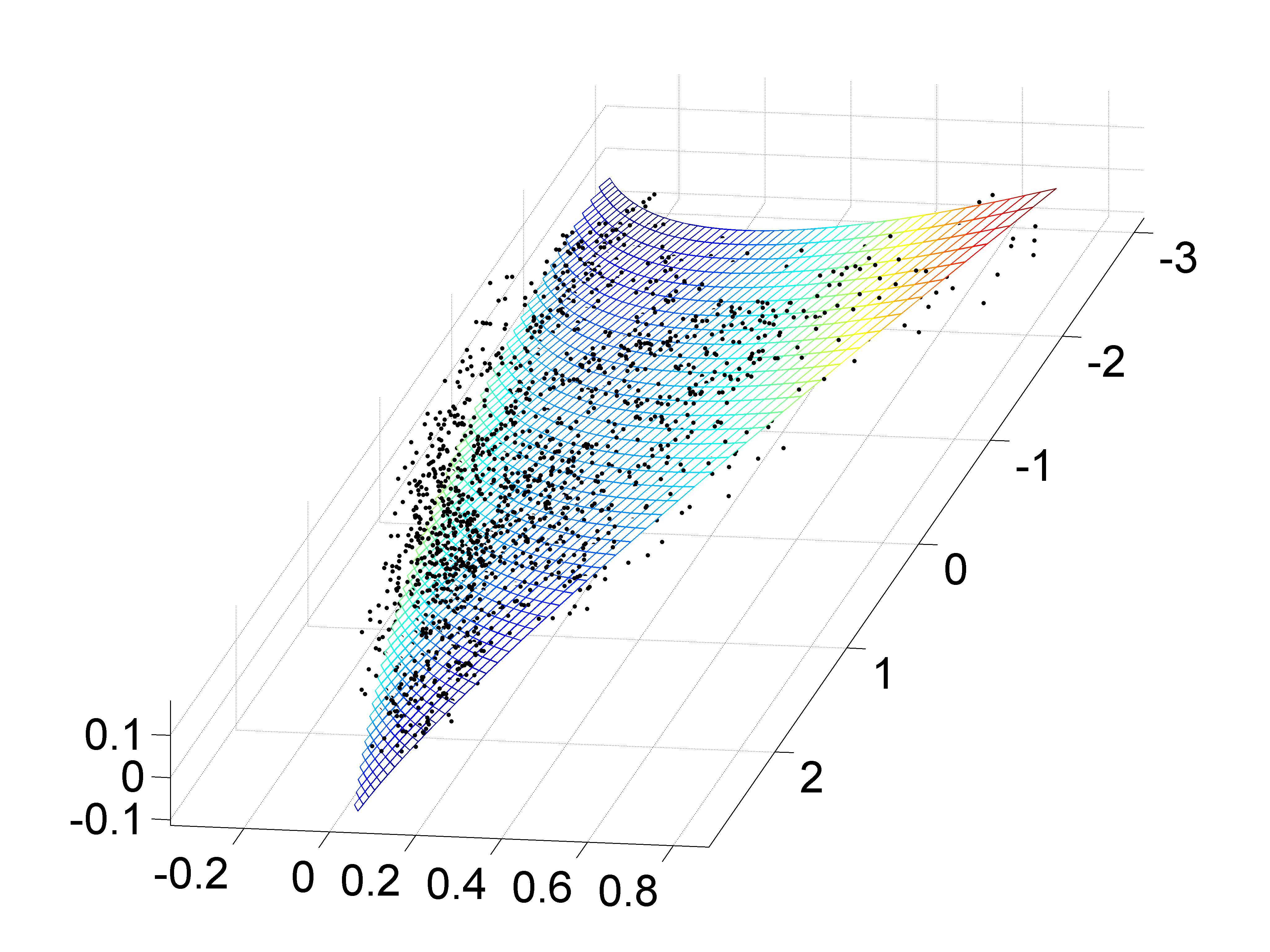}
  \centerline{(b)}\medskip
\end{minipage}
\hfill
\begin{minipage}[b]{0.48\linewidth}
  \centering
\includegraphics[width=\linewidth]{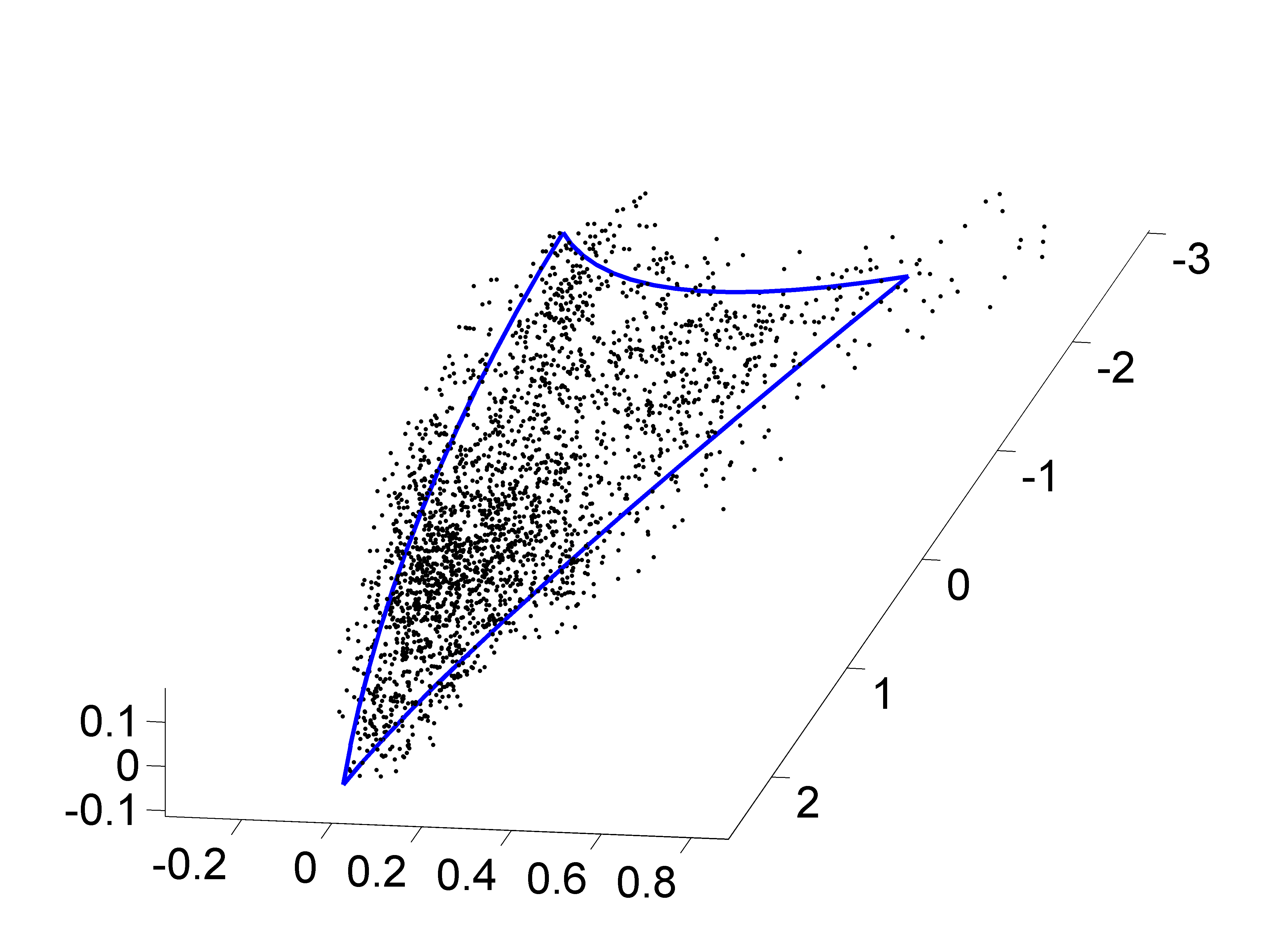}
  \centerline{(c)}\medskip
\end{minipage}
\vspace{-0.3cm} \caption{(a): Representation of the $N=2500$ pixels
(black dots) of the Madonna image and the ground
  truth clusters corresponding to oak trees (red dots) and chestnut trees (blue dots)
  using the first three principal components
  provided by the standard PCA. (b): Representation of the $N=2500$ pixels (dots) of the
  Madonna data and manifold identified by the LL-GPLVM (colored surface).
  (c):Representation of the $N=2500$ pixels (dots) of the Madonna data and boundaries of the estimated transformed simplex (blue lines).}
\label{fig:representation_real}
\end{figure}
\begin{figure}[h!]
  \centering
  \includegraphics[width=\columnwidth]{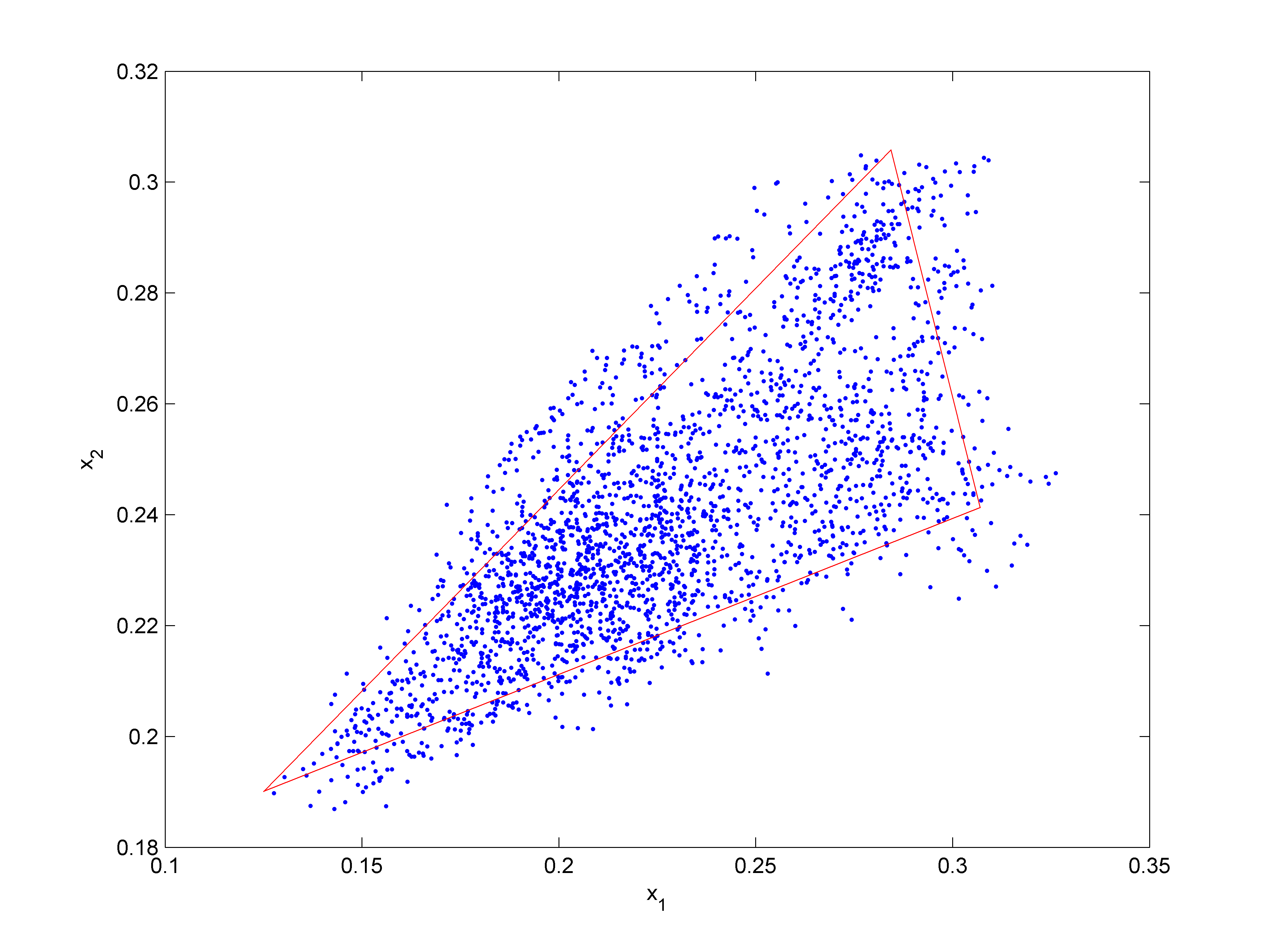}
  \caption{Representation of the $N=2500$ latent variables (dots) estimated by the LL-GPLVM and the simplex identified by the
  scaling step (red lines) for the Madonna data.}
\label{fig:Madonna_LV}
\end{figure}
\begin{figure}[h!]
  \centering
  \includegraphics[width=\columnwidth]{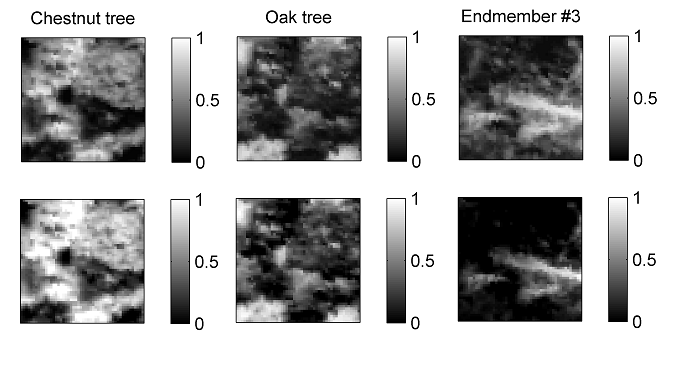}
  \caption{Top: Abundance maps estimated using the FCLL-GPLVM for the Madonna image. Bottom: Abundance maps
  estimated using the VCA algorithm for endmember extraction and the FCLS algorithm for abundance estimation.}
\label{fig:Madonna_abond}
\end{figure}

The real image considered in this section was acquired in 2010 by
the Hyspex hyperspectral scanner over Villelongue, France. $L=160$
spectral bands were recorded from the visible to near infrared with
a spatial resolution of $0.5$m. This dataset has already been
studied in \cite{Sheeren2011} and is mainly composed of different
vegetation species. The sub-image of size $50 \times 50$ pixels
chosen here to evaluate the proposed unmixing procedure is depicted
in Fig. \ref{fig:Madonna_big}. This image is mainly composed of
three components since the data belong to a two-dimensional manifold
(see black dots of Fig. \ref{fig:representation_real} (a)).
Consequently, we assume that the scene is composed of $R=3$
endmembers. Using the ground truth used in \cite{Sheeren2011}, we
can determine some tree species present in the scene of interest.
More precisely, Fig. \ref{fig:representation_real} (a) shows the
ground truth clusters corresponding to oak trees (red dots) and
chestnut trees (blue dots) projected in a 3-dimensional subspace
(defined by the first three principal components of a PCA applied to
the image of Fig. \ref{fig:Madonna_big}). These two clusters are
close to vertices of the data cloud. Consequently, oak and chestnut
trees are identified as endmembers present in the image. Moreover,
the third endmember is not a tree species according to the ground
truth provided in \cite{Sheeren2011}. In
the sequel, this endmember will be referred to as Endmember $\sharp3$.\\
The simulation parameters have been fixed to $\gamma=10^3$ and
$K=R$. The latent variables obtained by maximizing the marginalized
posterior distribution \eqref{eq:joint_posterior} are depicted in
Fig. \ref{fig:Madonna_LV} (blue dots).
It can be seen from this figure that the latent variables seem to
describe a noisy simplex. Fig. \ref{fig:representation_real} (b)
shows the manifold estimated by the proposed LL-GPLVM. This figure
illustrates the capacity of the LL-GPLVM for modeling the nonlinear
manifold. Table \ref{tab:ARE_real} (left) compares the AREs obtained
by the proposed LL-GPLVM and the projection onto the first $R-1=2$
principal vectors provided by PCA. The proposed LL-GPLVM slightly
outperforms PCA for the real data of interest, which shows that the
proposed nonlinear dimensionality reduction method is more accurate
than PCA (linear dimensionality reduction) in representing the data.
The scaling step presented in Section \ref{sec:scaling} is then
applied to the estimated latent variables. The estimated simplex
defined by the latent variables is depicted in Fig.
\ref{fig:Madonna_LV} (red lines). Fig. \ref{fig:representation_real}
(c) compares the boundaries of the estimated transformed simplex
with the image pixels. The abundance maps obtained after the scaling
step are shown in Fig. \ref{fig:Madonna_abond} (top). The results of
the unmixing procedure using the FCLL-GPLVM are compared to an
unmixing strategy assuming the LMM. More precisely, we use VCA to
extract the endmembers from the data and use the FLCS algorithm for
abundance estimation. The estimated abundance maps are depicted in
Fig. \ref{fig:Madonna_abond} (bottom). The abundance maps obtained
by the two methods are similar which shows the accuracy of the
proposed unmixing strategy when considering the LMM as a first order
approximation of the mixing model.
\begin{table}[h!]
\renewcommand{\arraystretch}{1.2}
\begin{footnotesize}
\begin{center}
\caption{AREs: real image ($\times 10^{-2}$).\label{tab:ARE_real}}
\begin{tabular}{|c|c||c|c|}
\hline
PCA & LL-GPLVM & VCA+FCLS & FCLL-GPLVM \\
\hline
0.84 & \textbf{\blue{0.79}} & 1.30 & \textbf{\blue{1.11}} \\
\hline
\end{tabular}
\end{center}
\end{footnotesize}
\vspace{-0.4cm}
\end{table}

\begin{figure}[h!]
  \centering
  \includegraphics[width=\columnwidth]{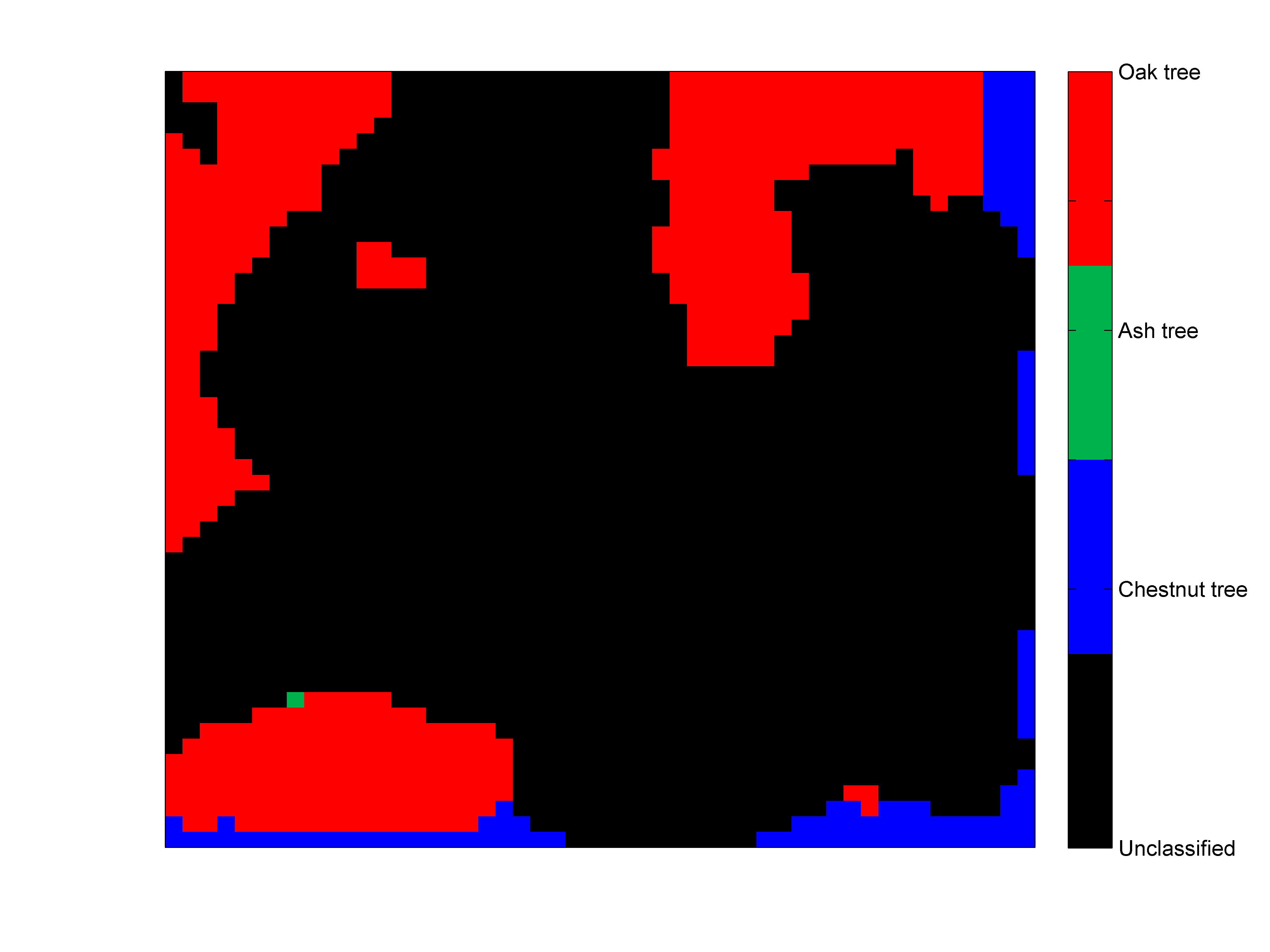}
  \caption{Classification map obtained in \cite{Sheeren2011} for the region of interest of the Madonna image.}
\label{fig:Madonna_classif}
\end{figure}

Moreover, Fig. \ref{fig:Madonna_classif} shows the classification
map obtained in \cite{Sheeren2011} for the region of interest. The
unclassified pixels correspond to areas where the classification
method of \cite{Sheeren2011} has not been performed. Even if lots of
pixels are not classified, the classified pixels can be compared
with the estimated abundance maps. First, we can note the presence
of the same tree species in the classification and abundance maps,
i.e., oak and chestnut. We can also see that the pixels composed of
chestnut trees and Endmember $\sharp3$ are mainly located in the
unclassified regions, which explains why they do not appear clearly
in the classification map. Only one pixel is classified as being
composed of ash trees in the region of interest. If unclassified
pixels also contain ash trees, they are either too few or too mixed
to be considered as mixtures of an additional endmember in the
image. Finally, it can be seen from Figs. \ref{fig:Madonna_abond}
and \ref{fig:Madonna_classif} that oak trees are located within
similar regions (left corners and top right corner) for the
abundance and classification maps.

Evaluating the performance of endmember estimation on real data is
an interesting problem. However, comparison of the estimated
endmembers with the ground truth is difficult here. First, since the
nature of Endmember $\sharp3$ is unknown, no ground truth is
available for this endmember. Second, because of the variability of
the ground truth spectra associated with each tree species, it is
difficult to show whether VCA or the proposed FCLL-GPLVM provides
the best endmember estimates. However, the AREs obtained for both
methods (Table \ref{tab:ARE_real}, right) show that the FCLL-GPLVM
fits the data better than the linear SU strategy, which confirms the
importance of the proposed algorithm for nonlinear spectral
unmixing.

\section{Conclusions}
\label{sec:conclusion} We proposed a new algorithm for nonlinear
spectral unmixing based on a Gaussian process latent variable model.
The unmixing procedure assumed a nonlinear mapping from the
abundance space to the observed pixels. It also considered the
physical constraints for the abundance vectors. The abundance
estimation was decomposed into two steps. Dimensionality reduction
was first achieved using latent variables. A scaling procedure was
then proposed to estimate the abundances. After estimating the
abundance vectors of the image, a new endmember estimator based on
Gaussian process regression was investigated. Simulations conducted
on synthetic and real images illustrated the flexibility of the
proposed model for linear and nonlinear spectral unmixing and
provided promising results for abundance and endmember estimations
in spite of the absence of pure pixels in the image. The choice of
the nonlinear mapping used for the GP model is an important issue to
ensure that the LL-GPLVM is general enough to handle different
nonlinearities. In particular, different mappings could be used for
intimate mixtures. Moreover, including the GPLVM in a Bayesian
framework could also be an interesting prospect to consider spatial
correlation through Markov random fields or GPs.

\section*{Acknowledgments}
The authors would like to thank Dr. Mathieu Fauvel from the
University of Toulouse - INP/ENSAT, Toulouse, France, for supplying
the real image, ground truth data and classification map related to
the classification algorithm studied in \cite{Sheeren2011} and used
in this paper.

\bibliographystyle{ieeetran}
\bibliography{biblio}

\end{document}